\documentclass[a4paper, twocolumn]{article}

\usepackage[a4paper,margin=16mm]{geometry}

\usepackage{amsmath,amsfonts}
\usepackage{algorithmic}
\usepackage{algorithm}
\usepackage{array}
\usepackage{subfigure}
\usepackage{textcomp}
\usepackage{stfloats}
\usepackage{url}
\usepackage{wasysym}
\usepackage{verbatim}
\usepackage{color}
\usepackage[table]{xcolor}
\usepackage{graphicx}
\usepackage{multirow}
\usepackage[hidelinks]{hyperref}
\usepackage{soul,xcolor}
\usepackage{tikz}
\usepackage{wrapfig}
\usepackage{booktabs}
\usepackage{threeparttable}
\usepackage[capitalise]{cleveref}
\usepackage{expl3,xparse} % to count cit
\usepackage{authblk}

\usepackage[acronym]{glossaries}
\usepackage[acronym]{glossaries}
\newacronym{cnns}{CNNs}{Convolutional Neural Networks}
\newacronym{gcnns}{GCNNs}{Graph Convolutional Neural Networks}
\newacronym{phri}{p-HRI}{phisical Human-Robot Interaction}
\newacronym{som}{SOM}{Self-Organizing Map}
\newacronym{gpis}{GPIS}{Gaussian Process Implicit Surfaces}
\newacronym{sdf}{SDF}{Signed Distance Functions}
\newacronym{slam}{SLAM}{Simultaneous Localization and Mapping}
\newacronym{vt_slam}{VT-SLAM}{Visuo-Tactile SLAM}
\newacronym{ml}{ML}{Machine Learning}
\newacronym{dl}{DL}{Deep Learning}
\newacronym{soa}{SoA}{state-of-the-art}
\newacronym{llms}{LLMs}{Large Language Models}
\newacronym{fms}{FMs}{Foundation Models}
\newacronym{vla}{VLA}{Vision-Language-Action}
\newacronym{eit}{EIT}{Electrical Impedance Tomography}
\newacronym{pps}{PPS}{Peripersonal Space}
\newacronym{tof}{ToF}{Time of Flight}

\usepackage{moreverb,url}

\newcommand\BibTeX{{\rmfamily B\kern-.05em \textsc{i\kern-.025em b}\kern-.08em
		T\kern-.1667em\lower.7ex\hbox{E}\kern-.125emX}}

\begin{document}

	\title{Representing Data in Robotic Tactile Perception - A Review}
	
	%%%%%%%%%% Authors Info %%%%%%%%%%%%%%%%%
	\author[1]{Alessandro Albini}
	\author[2,3]{Mohsen Kaboli}
	\author[4]{Giorgio Cannata}
	\author[1]{Perla Maiolino}
	\affil[1]{Are with the Oxford Robotics Institute (ORI), University of Oxford, United Kingdom.}
	\affil[2]{Is with the Eindhoven University of Technology (TU/e), Netherlands.}
	\affil[3]{Is with the AI, Robotics, \& Cognitive Vehicel Lab, BMW Group, Munich, Germany.}
	\affil[4]{Is with the Department of Informatics, Bioengineering, Robotics and Systems Engineering (DIBRIS), University of Genoa, Genoa, Italy.}
	%%%%%%%%%% End of Authors Info %%%%%%%%%%%%%%%%%
	
   \makeatletter
   \twocolumn[
   \begin{@twocolumnfalse}
   	\date{}
   	\maketitle
   	\begin{abstract}

%The sense of touch has been proven to be essential for robots operating in unstructured environments to ensure robustness and safety.
Robotic tactile perception is a complex process involving several computational steps performed at different levels. Tactile information is shaped by the interplay of robot actions, the mechanical properties of its body, and the software that processes the data.
In this respect, high-level computation, required to process and extract information, is commonly performed by adapting existing techniques from other domains, such as computer vision, which expects input data to be properly structured.

Therefore, it is necessary to transform tactile sensor data to match a specific data structure. This operation directly affects the tactile information encoded and, as a consequence, the task execution. 
This survey aims to address this specific aspect of the tactile perception pipeline, namely \textit{Data Representation}.

The paper first clearly defines its contributions to the perception pipeline and then reviews how previous studies have dealt with the problem of representing tactile information, investigating the relationships among hardware, representations, and high-level computation methods. The analysis has led to the identification of six structures commonly used in the literature to represent data. The manuscript provides discussions and guidelines for properly selecting a representation depending on operating conditions, including the available hardware, the tactile information required to be encoded, and the task at hand.
	
\end{abstract}
   	\vspace{1em} % small space before the two-column text (optional)
   \end{@twocolumnfalse}
   ]
   \makeatother
	
	\section{Introduction}

Autonomous robots operating in unstructured environments require the ability to interact with humans or objects. 
Tactile sensing plays a key role in supporting robots across diverse tasks, enabling the control of contact forces and the detection and management of collisions \cite{calandra_2015,killpack_2016,pugach_2016}.
Its significance in robotics has grown over time.
Initially confined to prototypes embedded into fingertips to support grasping and manipulation tasks \cite{howe_1990,melchiorri_2000,begej_1988}, tactile sensors have evolved into sophisticated systems capable of covering extensive portions of the robot's body and providing multimodal information \cite{ohmura_2006,cheung_1989,minato_2007,maiolino_2011,mittendorfer_2011}.
 
Tactile feedback provided by these sensors is the result of an implicit processing performed by the interplay between control actions and the morphological properties of the sensor. The control action directly contributes to shaping the resulting tactile information - pushing or rubbing against a surface aids the perception of different types of features. 
Aspects related to the morphology of the sensor, including the distribution of tactile receptors, as well as the shape and physical characteristics of the material of the surface that embeds the sensors, \textit{implicitly} filter physical stimuli, affecting the content of the tactile signal.
On top of this, to extract features of interest from tactile data, \textit{explicit} high-level processing and computation (at the software level) are applied to the sensor output. 
As a result, the whole interplay among the control action, hardware and high-level data processing contributes to tactile perception. 

Within this interplay of implicit and explicit computation, high-level reasoning is typically implemented using state-of-the-art algorithms that require input data to be properly structured.
It is then a common practice to represent tactile data provided by the hardware using structures compatible with algorithms performing high-level computations \cite{dahiya_2013,luo_2017,shih_2020}.
This aspect plays a crucial role, as the structure directly affects the encoded tactile information and, as a consequence, the task execution.
	Existing reviews already address other aspects of robotic tactile perception. Some primarily focus on sensing technologies and manufacturing processes \cite{natale_2016, baldini_2022, roberts_2021, yousef_2011, zhu_2022}. Others examine both technological and computational perspectives within the context of specific applications - such as the representation and recognition of object properties \cite{liu_2017, luo_2017}, manipulation tasks \cite{wei_2018, xia_2022}, and social human–robot interaction \cite{tawill_2014}.
	The review in \cite{qiang_2020}, instead, investigates the use of tactile sensing across a broader set of applications and focuses on the interplay between perception and action in computing and utilizing tactile information.
	Meanwhile, \cite{hu_2023} specifically focuses on the application of \gls{ml} methods for tactile perception, covering aspects of data acquisition, preprocessing, feature extraction, and classification.
	Finally, a number of surveys take a broad field-wide perspective, including \cite{luo_2025, dahiya_2010, dahiya_2013}, which provide system-level overviews of tactile sensing in robotics, covering sensor technologies, fabrication approaches, applications, and long-term challenges.

	In the context of tactile perception, data representation determines the tactile features that can be encoded, the contact properties captured, and the way data is interpreted and used.
	Despite its importance, this topic has often been overlooked, with prior studies focusing primarily on hardware or higher-level processing and computation. 
	As a result, key questions remain open on how tactile data should be represented depending on hardware and computational requirements.
	Compared to existing surveys, this manuscript addresses tactile perception from a complementary perspective. It fills the gap by providing an in-depth analysis of how tactile data can be structured and how this affects the perception pipeline.
	The manuscript begins with a general overview in Section 2, outlining factors that shape tactile information and clarifying where data representation fits within the overall tactile processing pipeline.
	Section 3 then examines its role as the bridge between hardware and high-level computational methods. Given that representation influences tactile perception at multiple levels, the following sections organize the discussion around four questions:
	\begin{itemize}
		\item \textit{How are state-of-the-art representations built and selected based on available hardware?}
		Sections 4 and 5 examine how tactile data can be structured despite hardware-specific characteristics. 
		\item \textit{What information is captured by the representation?} 
		Section 6 focuses on how the representation captures or filters tactile information. It highlights the types of tactile features each representation can encode and discusses the implications for their usage in higher-level computation.
		\item \textit{How does data representation support high-level computation in the perception pipeline?
		}
		Section 7 analyses how the representation supports robotic applications that leverage tactile feedback. It emphasizes the link between representation choice and task requirements, providing guidelines for application-driven selection of structures.
		\item \textit{How can distributed forces and pressure data be effectively represented alongside heterogeneous information from different sensors?}
		Section 8 extends the discussion to multimodal sensing, exploring how tactile data can be integrated with other sensing modalities. %It addresses the growing need for effective multimodal data fusion in robotics.
	\end{itemize}
	
Conclusions and discussions follow.

	\section{Tactile Perception in Robotic Systems}
\label{sec:overview}

\begin{figure*}[t!]
	\centering
	\includegraphics[width=0.99\textwidth]{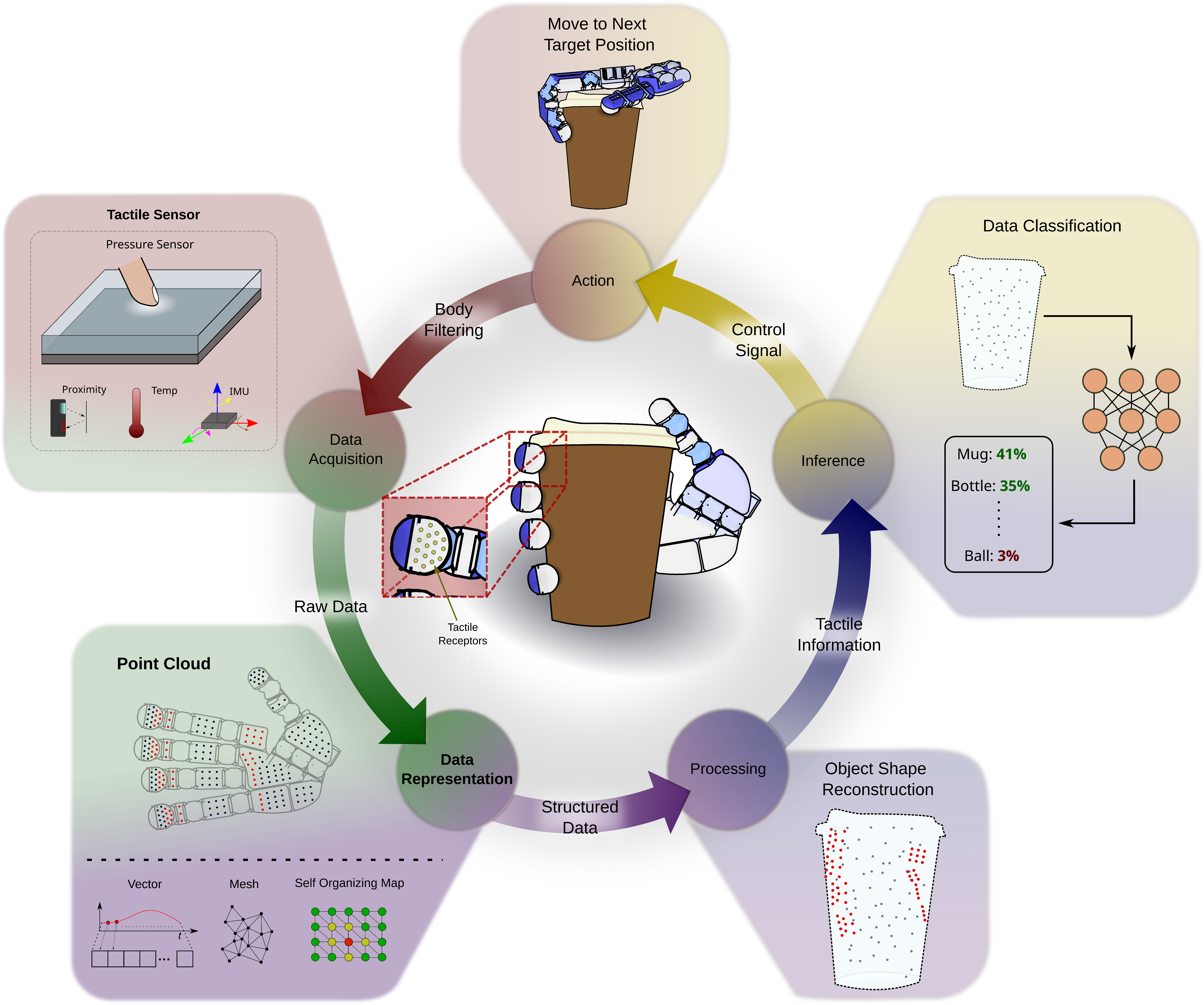} 
	\caption{
		General overview of the operations implementing a tactile perception pipeline.
		This review focuses on the Data Representation block highlighted in bold.
		Tactile perception is action-conditioned, thus creating the circular dependency among the five different operations shown in the figure.
		As an example, the image shows a possible content for each block in a task of tactile-based exploration using a robotic hand embedding tactile receptors.
		%.  
		The Data Acquisition block collects measurements from distributed sensors (previously filtered by the morphology of the body and robot actions) providing raw data as an output.
		The Data Representation block transforms the input to provide structured data that can be further processed at higher levels. The image shows some of the data structures used to represent data. In this example of tactile exploration the point cloud, representing the position of each receptor in space, is used. Red dots correspond to receptors in contact with the mug. 
		Structured data is then further elaborated to extract features of interest from the contact, or as shown in this case, to merge multi-contact information and create a point cloud representation of the object. 
		The inference step is related to high-level computation performing decision making. In this example, a data-driven model assigning a probability to each label, decides where to move at the next stage to increase the classification accuracy.
		Finally, the robot applies the Action, which affects the perceived information. In the example, the hand is moved to a target position to grasp the object with a desired force, closing the loop.
	}
	\label{img:overview}	
\end{figure*}

Tactile perception in robotics shares similarities with the biological counterpart. Humans extract tactile information as the result of different computational steps performed by both the body and the brain. The body plays a role in filtering physical contact. The skin's softness and structure work as a mechanical low-pass filter, spreading and modulating contact forces over the skin surface. Different areas of the body have varying densities of receptors, thus providing different acuity. 
Tactile receptors generate electrical signals transmitted through the spinal cord to the brain, which are organized and processed in the somatosensory cortex. 
Tactile information is then integrated with other sensory data to extract higher-level understanding of touch (e.g. material properties or the object's shape).
Through cognitive processes, the brain interprets and extracts meaning from tactile information and decides how to react  \cite{kandel_2000}. The subsequent physical action plays a crucial role in sensing and perception. Sense of touch does not passively receive information from the environment - the human body actively interacts with objects to extract meaningful data through active touch.

Robotic tactile perception systems follow similar processing and computational steps as their biological counterparts. \Cref{img:overview} illustrates this general pipeline, along with an example implementation for a tactile exploration task.
Due to embodiment aspects \cite{pfeifer_2007}, tactile perception is action-conditioned. Robot actions influence tactile data, which in turn guides subsequent actions, forming the circular dependency shown in \cref{img:overview}, where different stages are described in the following:
\begin{itemize}
	\item \textbf{Action:} Tactile data is generated as a result of a robot's actions, typically involving movements that create physical interactions with the environment. The nature of tactile information depends on the type of action applied. For example, by rubbing against a surface, high-frequency tactile information is generated. These patterns are influenced by the speed of the movement, the applied force, and the mechanical properties of the surface. This can be useful for discriminating materials based on their roughness or texture \cite{kerr_2018}. In contrast, applying a static force generates a pressure distribution whose intensity and area can be adjusted by actively regulating the contact force, allowing the retrieval of geometric information
	\cite{berger_1991}.
	\item \textbf{Data Acquisition:} Tactile sensors distributed on the robot body capture the effects of physical interactions between the robot and the external world, producing raw data as output. The format and content of raw data are discussed in more detail in Section 4. %\cref{sec:hardware}.
	At this stage a level of filtering is applied by the morphology of the sensor - its mechanics acts as a filter, shaping the information provided. For example, its softness affects the resulting contact area. The captured information also depends on the distribution of receptors, their spatial resolution, and both size and geometry of the robot surface \cite{dahiya_2010}.
	Furthermore, depending on the type of transducer embedded in the sensor, different effects of the physical interaction can be captured, such as vibrations or temperature variations. This type of filtering is conceptually similar to the one applied implicitly by the body in human tactile perception.
	In addition, for artificial systems, the overall electronics for data transmission and the firmware set limits on the sampling time and the data resolution, further filtering the original contact information.
	\item \textbf{Data Representation:} Raw data acquired by the hardware are then transformed or re-arranged to be represented by means of data structures supporting further levels of processing. 
	It is worth noting that, from the biological point of view there is a weak relation to processes in human tactile perception. Signals provided by mechanoreceptors are organized and mapped into the somatotopic map contained in the somatosensory cortex \cite{kandel_2000}. Similarly, structures used to represent artificial tactile information can associate raw data with the different areas of the robot body \cite{hoffmann_2018}.
	However, beyond this analogy, data representation in robotics is mainly needed both from the system implementation and integration points of view. The restructuring of data performed at this stage allows for interfacing raw data with state-of-the-art processing algorithms, despite the differences in the sensing hardware (more details are given in Section 3). % \cref{sec:middleware}).
	The output of this block consists of well-structured data, highlighting specific aspects of the contact information. 
	\item \textbf{Processing:} Structured data is processed to extract basic contact information, such as contact parameters, or to highlight or extract features of interest for the task. Examples of processing operations include simple thresholding, computation of contact locations or merging of multi-contact information.
	In the example shown in \cref{img:overview}, data represented as point cloud and collected from multiple contacts are merged to reconstruct the shape of the object.
	\item \textbf{Inference:} At this stage, processed tactile information is interpreted to make a decision on the next action of the robot. Examples of inference operations include the detection of edges, object recognition and slip detection. 
	The outcome of this operation is used to properly generate a control signal for the robot corresponding to a physical action, thus closing the loop in \cref{img:overview}. In the example, the inference step consists of a classifier that recognizes the object's point cloud and commands the robot to a grasping pose that improves classification accuracy.
\end{itemize}
It is important to note that \cref{img:overview} shows the conceptual architecture. 
In this respect, two aspects must be highlighted.
First, the three blocks Processing, Inference, and Action may not be explicitly separated from an implementation point of view. 
Instead, end-to-end approaches, particularly those based on deep learning, can integrate these stages into a single architecture, enabling a direct mapping from structured tactile data to robot actions.
For instance, as shown in \cite{lee_2020}, a single neural network can be trained to perform all three steps jointly.
Second, data representation is not necessarily a fixed, one-time transformation. As tactile data flows through the stages in \cref{img:overview}, it may need to be restructured to suit the processing requirements at each stage. Consequently, the representation layer does not maintain a single fixed data structure throughout the entire task execution. Instead, it can contain multiple representations of the same underlying tactile data tailored to task-specific requirements.

	\section{Representing Tactile Information}
\label{sec:middleware}

The contribution of the Data Representation stage in \cref{img:overview} can be evaluated from two complementary perspectives:
	\begin{itemize}
		\item \textbf{Hardware Perspective} -
		Processing methods may not be directly compatible with the distribution of tactile receptors embedded in the robot body, which can be arranged in non-regular and/or non-planar configurations, with a density that varies across different body areas. In this context, Data Representation serves as a compatibility layer between processing algorithms and the specific morphology of the tactile sensor. 
		\item \textbf{Computational Perspective} - It enables the use of state-of-the-art processing methods not originally developed for tactile sensors.
		As an example, representing data in the form of images %\cite{schneider_2009,pezzementi_2011,liu_2012} 
		or meshes %\cite{fan_2022,albini_2021} % or point clouds \cite{falco_2019,parsons_2022},  
		enables further computation based on \gls{cnns} or \gls{gcnns} \cite{liu_2017,fan_2022}. 
		Depending on the task, different computational techniques can be considered, each expecting properly structured input data.
	\end{itemize}

The first aspect is mainly due to technical reasons. While it would be desirable to achieve the highest possible spatial resolution, uniform density, and high sampling frequency, these goals are constrained by technological limitations \cite{dahiya_2010}. Raising both spatial resolution and frequency increases bandwidth requirements, thereby limiting the area that can be sensorized.
In this respect, robotics fingertips may integrate high-spatial resolution and high-frequency sensors for tasks related to grasping and manipulation.
In contrast, larger surfaces, such as robot hands, arms, or the entire body, face scalability issues necessitating a trade-off between spatial resolution and frequency. 
More importantly, integrating sensors into different robot body areas may require different design principles to ensure that the sensor’s shape conforms to the specific part, making the distribution of the tactile receptors ill-defined \cite{maiolino_2011,dahiya_2013}.

Drawing a comparison with computer vision's image processing domain, the specific camera technology becomes irrelevant, as data is consistently presented as images that can be accessed and processed using standard techniques.
This is not possible in tactile data processing, and
to address this challenge, it becomes critical to adopt an approach based on \textbf{hardware abstraction}. By rearranging data into common, standardized structures, high-level processing methods can be decoupled from the underlying hardware configuration, thus avoiding the use of technology-dependent algorithms. % that would not be appealing from an engineering point of view. 
Hardware abstraction also favors a 
seamless integration of tactile feedback with data collected from different sensors, such as proximity sensors, cameras, or accelerometers. For instance, visuo-tactile data can be represented with the same format of point clouds or images \cite{murali_2021,nguyen_2021,luo_2018,zhao_2023}.

	 It is also important to note that Data Representation in the context of the tactile perception loop in \cref{img:overview}, should not be blurred or confused with object or task representations. These concepts are substantially different, and it is worth explicitly highlighting their distinctions.
	 While Data Representation focuses on how raw tactile signals are restructured, object and task representations serve distinct purposes:
	\begin{itemize}
		\item \textbf{Object/Environment Representation} - This refers to how tactile data, potentially combined with other sensing modalities, is used to encode and represent the physical properties and geometry of objects or the environment surrounding the robot. With respect to \cref{img:overview}, this typically falls within the Processing block. For example, using point clouds to represent an object’s shape is a form of object representation. Such representations are primarily employed for tasks like object recognition, estimation of physical properties, localization, and geometric modeling.
		\item \textbf{Task Representation} - It refers to how the goal, context, and required actions are modelled \cite{dutta_2025}.
		In the context of tactile sensing, it involves the encoding of dynamic, context-specific information needed to execute and adapt manipulation or control tasks, such as grasp stability or slip detection, to support decision-making and control during physical interaction. In relation to \cref{img:overview}, this is typically associated with the Inference or Action stages. 
	\end{itemize}
	In this respect, Data Representation operates at a lower level by providing the input on which object- and task-level representations are built. By determining how raw tactile measurements are organized and encoded, it constrains and affects the higher-level abstractions used for objects or tasks.

\section{Tactile Sensing Hardware Abstraction}

\label{sec:hardware}

\begin{figure*}[t!]
	\centering
	\includegraphics[width=0.98\textwidth]{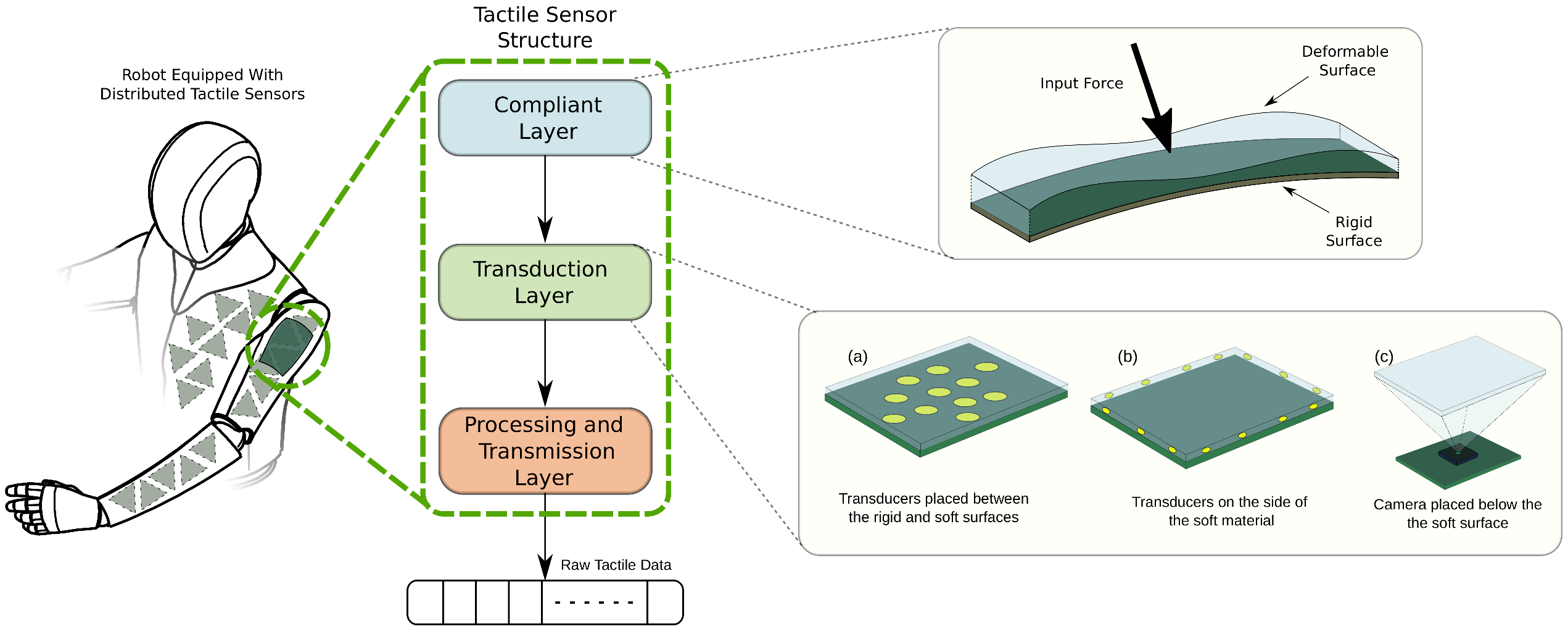} 
	\caption{A robot equipped with distributed tactile sensors represented as triangles. The image shows a high-level overview of the three layers composing a tactile sensor. The compliant layer is made of a material that deforms when a force is applied to it. The deformation is then captured by transducers. The force measurements are then collected, organized and transmitted as raw data.}
	\label{img:hardware}
\end{figure*}

This section explains how tactile sensing hardware can be modeled so that the Data Representation layer remains abstracted from technology-specific aspects. 
The discussion begins with a high-level overview of the main tactile sensing technologies and then introduces a set of concepts that make it possible to model the hardware in a unified way, despite differences in sensing principles and implementations.

\cref{img:hardware} illustrates a robot equipped with distributed tactile sensors. The robot's action (or a physical action applied to the robot) produces  \textit{force} or \textit{pressure distribution} measured by sensors through the deformation of a compliant structure. 
Although tactile sensors can be designed to provide multi-modal information on the contact event, this section considers sensors measuring force or pressure distributions, as they are the most commonly used in the literature. 
Additional modalities such as perception of temperature or vibrations are considered later in Section 8. % \cref{sec:multimodal}.
The structure of a tactile sensor can be summarized as in \cref{img:hardware}, composed of three macro layers, each filtering and affecting tactile information.
A proper design of the three layers shown in \cref{img:hardware} leads to the development of a device that captures desired properties of the original contact information.

\begin{enumerate}
	\item \textit{Compliant Layer}: A physical interaction applied on the surface causes a deformation. 
	The compliance of this soft medium determines the extent to which it conforms to the shape of the object in contact with the surface. This layer acts as a spatial filter, whose mechanical properties affect the overall response of the sensor including its sensitivity and static/dynamic range \cite{maiolino_2015}. 
	In this paper, we refer to the deformation assumed by the compliant layer as the \textbf{contact shape}.
	\item \textit{Transduction Layer}: \label{enum:trans_layer} The force distribution applied on the surface is captured indirectly by measuring the deformation of the soft medium using a set of transducers, for example capacitors \cite{maiolino_2011,mittendorfer_2011}, hall-effect sensors \cite{wang_tri_2016,tomo_2018}, resistive sensors \cite{weiss_2005,zhu_2022}, cameras \cite{lambeta_2020,shimonomura_2019} or electrodes \cite{tawill_2011,pugach_2016}. A complete list can be found in \cite{dahiya_2010,natale_2016}. The choice of the transducers affects the overall static/dynamic range of the measurement, as well as the type of information captured by the sensor (e.g. normal or shear forces).
	Depending on the technology, these transducers can be placed on top, below, embedded into the compliant layer \cite{ohmura_2006,weiss_2005,wang_2014,wang_tri_2016,maiolino_2011,lambeta_2020} or at its side \cite{tawill_2011,duong_2021}. 
	A schematic example of the possible placement of the transducers is given in \cref{img:hardware}; (a)-(b) illustrate hardware utilizing independent transducers, while (c) illustrates a camera-based tactile sensor capturing deformations by detecting changes in internal lighting or markers' movements within the medium \cite{shah_2021}.	
	\item \textit{Processing and Transmission Layer}:
	At this level, low-level processing is applied to the measurements. This layer deals with the sampling of the measurement, quantization and signal conditioning. Furthermore, it can synchronize the measurements coming from distributed transducers, and organize data to be transmitted over a bus. The output of this layer is the raw tactile data  corresponding to images for camera-based sensors \cite{lambeta_2020,yuan_2017,donlon_2018}, and to arrays of measurements for tactile technologies composed of independent transducers \cite{maiolino_2011,tawill_2011,pugach_2016}. 
	The design of this layer impacts the frequency content of the tactile signal, as well as the latency of the overall system.
\end{enumerate}

	To discuss data representations in a way that is independent of the specific hardware technology used, the concept of \textbf{tactile element} is introduced. 
	As illustrated in \cref{img:hardware}(a), traditional tactile sensors typically consist of discrete transducers embedded in or covered by a soft material. The spatial arrangement of these transducers approximates the \textbf{geometry} of the rigid surface (i.e., the robot body part), and, when combined with their individual readings, provides a spatial sampling of the external force distribution applied to the compliant layer.
	Within the tactile sensing community, these transducers are commonly referred to as tactile elements or \textbf{taxels}. Examples of taxel-based tactile sensors have been reported in the literature~\cite{schmitz_2011,ohmura_2006,mukai_2008,mittendorfer_2011,jamone_2015,paulino_2017}, and similar devices are also commercially available and have been used in recent studies~\cite{murali_2023,kitouni_2024}.

	This concept generalizes to other tactile sensing technologies, as shown in \cref{img:hardware}(b) and (c).
	In the case depicted in \cref{img:hardware}(b), transducers are instead placed along the sides of the soft material to obtain a continuous representation of the applied force distribution. The definition of taxel can still be applied, provided that the surface geometry is known. Examples of such technologies include those based on \gls{eit} and optical waveguide principles, where the geometry can be resampled. In this context, a taxel can be defined for each resampling point, as demonstrated in \cite{tawill_2011,bacher_2024}.
	For vision-based tactile sensors, a taxel can correspond to a single camera pixel or a tracked marker. 
	By grounding camera-based sensing in the taxel concept, the subsequent analyses can be applied uniformly across all tactile technologies. 

Furthermore, the concept of taxels can be extended to abstract transducers providing different information, such as the temperature or vibrations.
From this point on, the tactile sensing hardware is referred to as composed of taxels.

%\newpage

\section{Data Structures Representing Tactile Information}

\label{sec:build}

\begin{table*}[t]
	\centering
	\caption{
		Structures commonly used in the literature to represent tactile data. Some are built from raw taxel measurements alone, while others incorporate information on their spatial distribution.
	}
	\resizebox{\textwidth}{!}{
	\label{tb:tactile_structures}
	\begin{tabular}{@{}lll@{}}
		\toprule
		\textbf{Structure} & \textbf{Required Taxels' Information} & \textbf{Description} \\ \midrule
		Vector              & Raw measurements    & Flat container of raw sensor measurements; \\
		Matrix             & Raw measurements    & 2D grid-like container of raw sensors measurements. Indexing does not reflect spatial distribution. \\
		Map                & Raw measurements + \textbf{(optional)} Spatial Distribution & Abstract representation of the taxels' distribution. \\
		Point Cloud        & Raw measurements + Spatial Distribution & 3D coordinates of taxels in space. \\
		Mesh               & Raw measurements + Spatial Distribution &  Piece-wise approximation of the (non-planar) taxel distribution. \\
		Image              & Raw measurements + Spatial Distribution     & 2D grid where local neighbouring among taxels is preserved at best. \\ \bottomrule
	\end{tabular}
	}
\end{table*}

Section~3 introduced the role of Data Representation as the abstraction layer interfacing hardware and software, while Section~4 described how this abstraction can be derived from the hardware perspective through the generalized concept of taxel. Building on this, the current section discusses how representations are implemented in practice through concrete \textit{data structures} and explains how these can be created from different taxel distributions.

Data structures can be categorized into two groups as shown in \cref{tb:tactile_structures}. They can be built either from raw data measurements alone, or including the information of the taxels' distribution.
A discussion related to how a representation should be chosen depending on the hardware follows at the end of the Section.

\subsection{Data Structures Built from Taxels' Raw Measurements}

\subsubsection*{Vector}
A 1-dimensional array encoding raw measurements. 
It can be used to represent time series of data - each taxel is associated with a vector, whose elements correspond to measurements collected in a given time window.
Therefore, the vector captures the evolution of the single taxel measurement over time. This representation usually supports processing stages aimed at extracting features by considering a fixed time window of contact events.
The length of the time series and their temporal resolution are generally chosen according to the sampling time provided by the hardware or task requirements \cite{khan_2016, kerr_2018,martinez_2013,naya_1999,stiehl_2005,huisman_2013}. 

\subsubsection*{Matrix}

\begin{figure}[t!]
	\centering
	\includegraphics[width=0.48\textwidth]{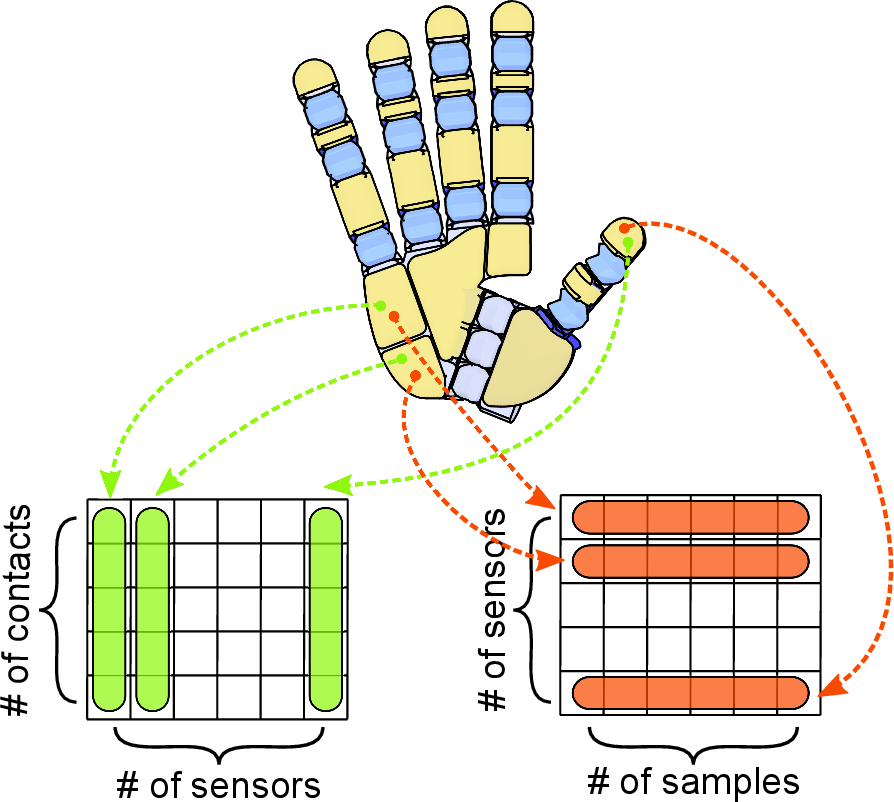}
	\caption{Two different ways exploited to rearrange raw data into a matrix structure. Yellow pads correspond to taxels distributed all over the hand. Using this representation \gls{ml} methods can be used to process data acquired in different contact positions or to extract time-dependent features. (Left) Raw data acquired from different contacts are stacked over the rows. (Right) The time series acquired from a single sensor is encoded as a row in the matrix. 
	}
	\label{img:matrix}
\end{figure}

A 2-dimensional array, mostly used as a container to leverage the use of \gls{ml} algorithms.
Raw data provided by each taxel are reorganised depending on the type of correlations that need to be highlighted. 
In \cite{shorthose_2023} data are rearranged as shown in the left part of \cref{img:matrix} - each row of the matrix corresponds to the response of each taxel captured during a grasp in a steady-state condition. Additional rows are stacked to embed information on the measurements acquired at different grasping postures. This allows convolutional kernels of \gls{cnns} to perform a feature extraction by considering %both adjacent columns and the 
raw data obtained at different grasping positions.
The work in \cite{gao_2015} proposes a reordering of the raw data to encode temporal information. % of the tactile response. 
The right part of \cref{img:matrix} shows that the matrix is built by associating each row with the response of a single taxel considered over a fixed-size time window. Columns instead represent the number of samples.
Convolutional layers will extract features related to a small time window corresponding to the width of the kernel. A similar approach was used in \cite{baishya_2016,wang_2022} to reshape the tactile data signal to allow the extraction of temporal-dependent features.

\subsubsection*{Map}
\label{subsec:map_raw}

\begin{figure}[t!]
	\centering
	\includegraphics[width=0.48\textwidth]{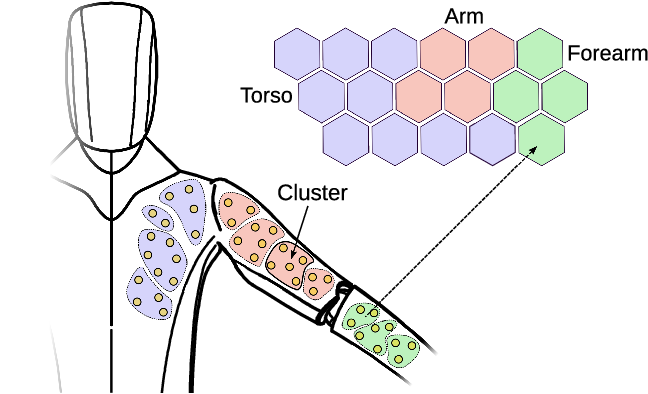}
	\caption{
		A map representation conceptually similar to the one proposed by \cite{hoffmann_2018}. Groups of taxels are stimulated to learn their spatial adjacency. Each hexagon in the map corresponds to a cluster position in 2D, with an associated value derived from the taxels' responses. The layout preserves proximity relations among clusters and across different body parts.
	}
	\label{img:map}
\end{figure}
This is used to build information on the taxels' distribution from the knowledge of their response alone. 
It can be created through a training procedure correlating taxels' measurements with physical actions. 
The result of this learning process maps the original spatial distribution into a planar space \cite{kuniyoshi_2004,brock_2009,pugach_2015,hoffmann_2018}. Data can be represented either as a set of points scattered in the plane or as graph-like structures. % a set of points or meshes in the 2D space.
As an example, a \gls{som} was considered by \cite{pugach_2015} to convert the raw array of responses into a planar representation resembling the topography of the sensor.
Different approaches proposed methods to group taxels into clusters that are semantically correlated \cite{hoffmann_2018,kuniyoshi_2004,brock_2009}. These representations are usually referred to as \textit{somatosensory maps}, where taxels belonging to the same robot part (e.g. head, arm or torso) are mapped close together in the planar representation. 
This type of representation mimics how humans organize and correlate sensory-motor stimuli in the brain, where the different parts of the somatosensory cortex are related to different body parts \cite{kandel_2000}. In robotic tasks, this representation is useful to learn how actions affect the output of the sensors and to build correlations among nearby taxels without explicitly knowing their location in space. A visual representation of this type of structuring, inspired from \cite{hoffmann_2018}, is illustrated in \cref{img:map}.

\subsection{Data Structures Built from the Taxels' Distribution}

\begin{figure*}[t!]
	\centering
	\includegraphics[width=0.98\textwidth]{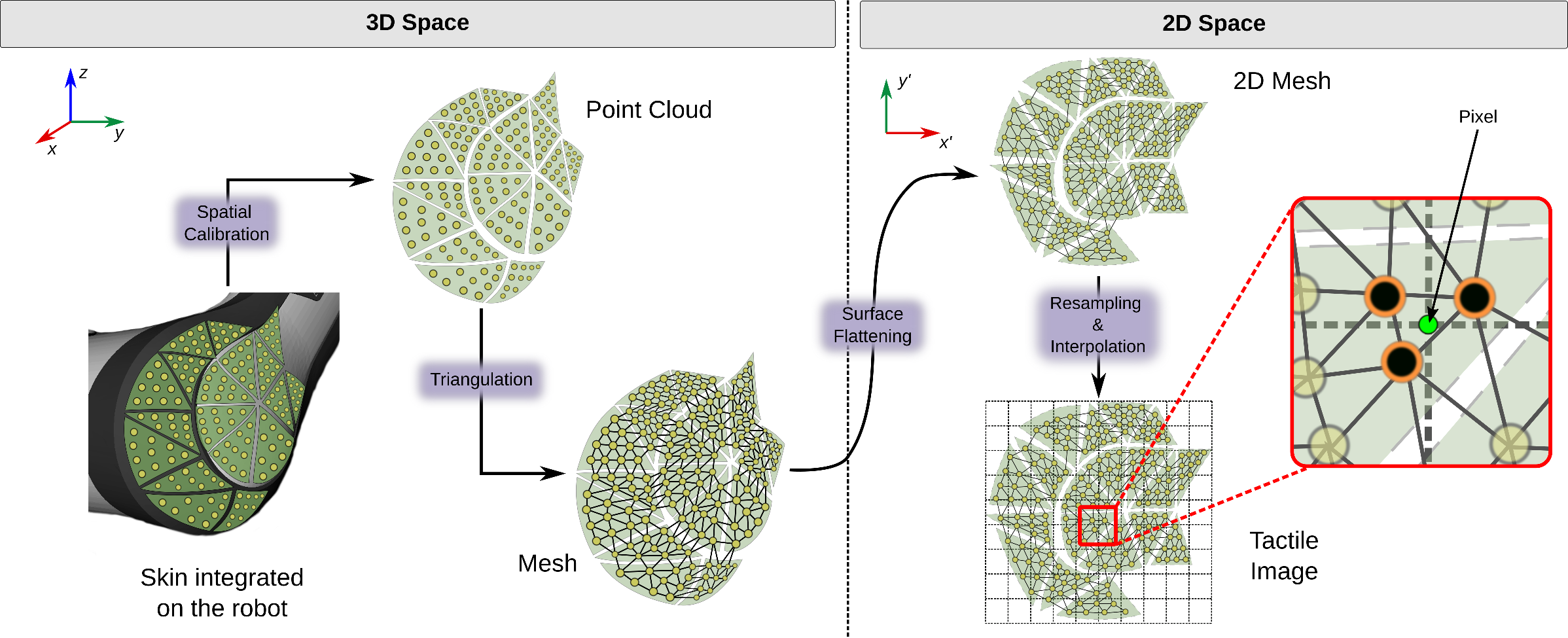}
	\caption{Steps to encode tactile data into point clouds, meshes and images starting from a non-regular and non-planar arrangement of the taxels. If taxels are spatially calibrated, their distribution in the space can be represented as a point cloud that can be further processed to obtain a mesh. The same mesh can then be flattened to lie on a plane and  resampled with a regular grid. The pixels' values in the image are computed by interpolation considering the measurements of the three adjacent taxels highlighted in orange.
 }
	\label{img:layout2representation}
\end{figure*}

\begin{figure}[t!]
	\centering
	\subfigure[]{\includegraphics[width=0.21\textwidth]{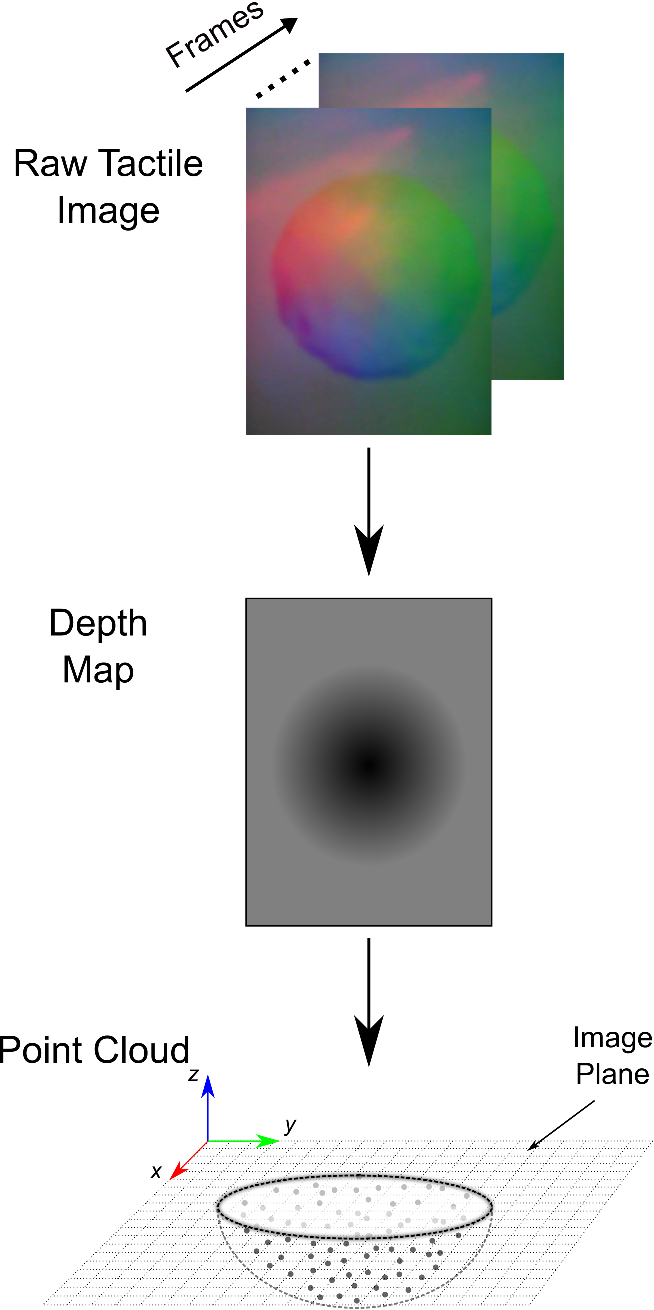} \label{img:rgb_build}} \quad
	\subfigure[]{\includegraphics[width=0.21\textwidth]{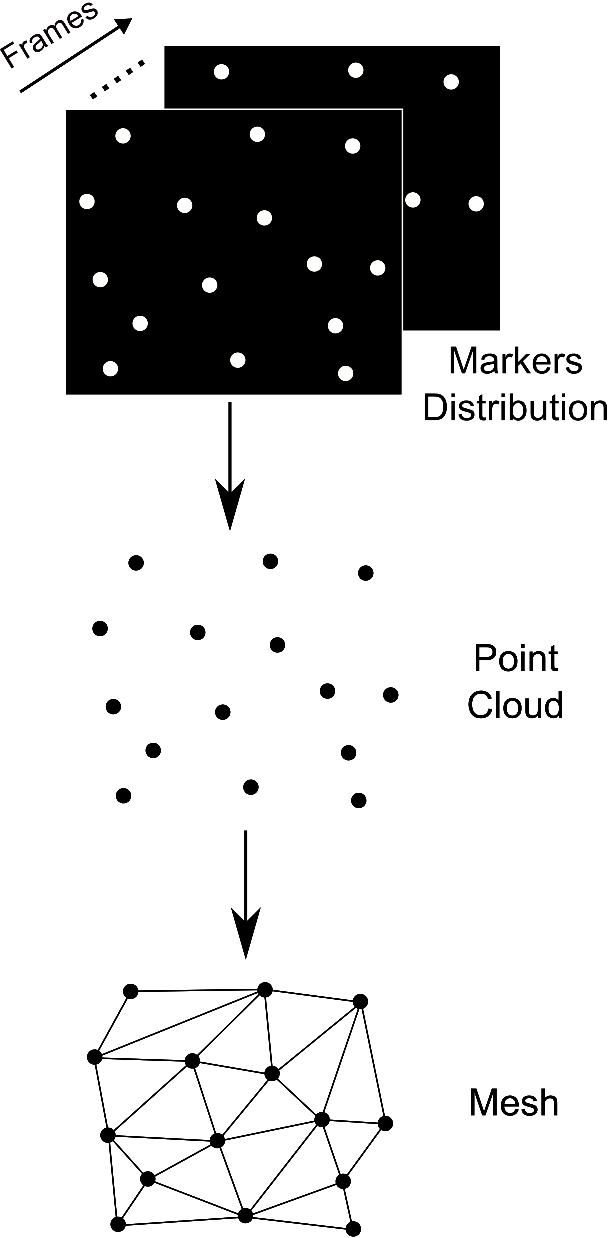} \label{img:marker_build}}
	\caption{Point cloud or meshes built from a camera-based tactile sensor. (a) The raw tactile image is converted into a depth map and successively into a point cloud representing the local deformation of the surface \cite{bauza_2019}. (b) Tactile images, encoding marker deformation, can be processed to extract their locations and convert them into a point cloud or mesh as shown in \cite{fan_2022,duong_2021,quan_2023}.}
	\label{img:build_from_image}
\end{figure}

The structures introduced in the following require a priori knowledge of the taxel distribution, either their adjacency or the exact position and orientation in space. 
In the latter case, taxels are defined to be \textit{spatially calibrated} \cite{cannata_2010}. 

\subsubsection*{Point Cloud}

It can be built from spatially calibrated taxels, such that there is a 1-to-1 correspondence between points and taxel locations.  
Thanks to their nature, point clouds can be used to encode both sparse and regularly arranged data and thus can be applied to represent taxels with any spatial distribution. As shown in the left part of \cref{img:layout2representation}, the point cloud represents an approximation of the surface where taxels are embedded.
Additionally, a value can be associated with each point, representing the measurement of the taxel related to the local deformation of the compliant medium, thus allowing the retrieval of information on the force distribution applied over the contact area.
They can also be created from raw tactile images, generated by camera-based sensors, in two different ways. The first is illustrated in \cref{img:rgb_build} and is analogous to having taxels following a grid-like pattern. A depth map is computed and converted into a small point cloud approximating the contact shape \cite{bauza_2019,suresh_2021_1}. The second is shown in \cref{img:marker_build} where the image frame is processed to extract the position of the markers (equivalent to taxels) which are then encoded as a set of points \cite{fan_2022,duong_2021,quan_2023}. 

\subsubsection*{Mesh}
It can be built by triangulating a set of points as explained in \cite{albini_2021,albini_2020,denei_2014} and illustrated in \cref{img:layout2representation}. 
Compared to point clouds, the proximity relationships among taxels are explicitly defined in the triangulation, thus providing a piece-wise approximation of the taxels' spatial distribution.
For camera-based tactile sensors capturing the motion of markers, the process is slightly different and follows the procedure in \cref{img:marker_build}. As shown in \cite{fan_2022,duong_2021,quan_2023}, images are first processed to extract the position of the markers and encode them as nodes. A triangulation is then applied on top of them. 
Similarly to point clouds, meshes can naturally manage sparse data, making them suitable to represent taxels having any spatial distribution.
From the point of view of the information captured, a mesh is equivalent to a point cloud. The choice between the two usually depends on the input type required by the algorithm performing high-level computation. 
As an example, point clouds can be considered when the contact location is required or to leverage the use of spatial descriptors or \gls{dl} architectures \cite{parsons_2022, falco_2019, killpack_2016, frigola_2006}.
By contrast, a mesh is worth considering when neighbouring relations among taxels need to be processed \cite{albini_2021} or when the use of \gls{gcnns} is required \cite{fan_2022}.

\subsubsection*{Image}

The value of each pixel composing the image is associated with the local deformation applied to the sensor. 
Therefore, the geometry of the object in contact with the compliant layer is mapped into a grid structure. % 
Tactile images can be defined from the knowledge of taxels' adjacencies when following a grid-like distribution, which may be either flat \cite{pezzementi_2011, luo_2015, liu_2012}, or where its curvature is not considered \cite{choi_2022}. 
For camera-based sensors, tactile images correspond to raw data, since the representation is already embedded in the data provided by the processing and transmission layer (discussed in Section 4), % \cref{sec:hardware}),
arranging the measurements into a grid.
When the spatial distribution is planar but not resembling a grid, data can still be transformed into images by resampling as shown in \cite{wasko_2019,muscari_2013}. A regular grid is superimposed onto the real taxels' placement and the value of each pixel can be computed by interpolation. 
If the spatial distribution is not only sparse but also non-planar, it can be flattened and constrained to lie in a plane \cite{albini_2020,taunyazov_2019} and then resampled as previously described. 
\cref{img:layout2representation} shows the general process when applied to taxels having non-planar and non-regular distribution.
Similarly to point clouds or meshes, tactile images can be used to capture information on the pressure distribution. The clear advantage of using this representation is related to the possibility of using all the existing methods developed for image processing.
It is also worth noting that multiple tactile images, built at different time instants, can be stacked as additional channels to encode the evolution of the spatial distribution over time, as shown by \cite{choi_2022}.

\subsubsection*{Map}
\label{subsec:map_calib}
Beyond the previous introduction in Section 5.1,
%\cref{subsec:map_raw}, 
a map can also be built from a spatially calibrated sensor as shown in \cite{sutanto_2019,denei_2014}.
The non-planar taxels' distribution is transformed and mapped into a plane using a flattening algorithm or an encoder/decoder network \cite{denei_2014,sutanto_2019}.  
The corresponding planar placement of the taxels does not necessarily approximate the original one, but the representation is used to perform computations which may be complex in 3D domain. 

\subsection{Discussion}

The data structures introduced above are built by organising or transforming the measurements and/or spatial distribution of the taxels. 
To select them depending on the hardware layout,
two aspects related to the spatial distribution should be considered:
\begin{itemize}
	\item \textbf{Complexity}: 
		Although the structures introduced in Section 5.2 explicitly or implicitly encode the layout of the taxels, they can be challenging to build in practice.
		Spatial calibration of taxels becomes particularly difficult when the tactile sensor has a complex, non-planar geometry and a high density of taxel distributed over a large area.
		This remains an open research problem, and a general, autonomous calibration solution that works across different types of sensing hardware has yet to be established.
		While a detailed discussion of this issue lies beyond the scope of this section, additional explanations and considerations are provided in the Appendix.
	In such cases, representations like vectors, matrices, and maps are often preferable, as they can be constructed directly from raw measurements without requiring spatial calibration.
	\item \textbf{Pattern}: taxels having a grid-like pattern can easily be represented using tactile images. 
	Point clouds or meshes are more suitable for non grid-like distributions as they can encode sparse data.
	Although images can still be applied by performing a resampling and interpolation (in addition to a flattening in the non-planar case), this transformation may result in loss of information. This aspect is investigated in detail in the next Section. 
\end{itemize}
These considerations illustrate how the spatial layout of the sensing hardware can guide the selection of the structure. However, it does not account for task-level requirements. In general, a trade-off should be found.
As an example, representations built from raw data are generally simpler to build but suffer from the drawback of not being suitable for various tasks that require geometric information, such as the precise locations of contacts. 
For what concerns the taxels' arrangement, it would be more immediate to use a data structure matching the taxels' spatial distribution. However, the task may strictly require a different representation. 
For instance, despite the hardware, the task may need data represented as an image to leverage the use of \gls{cnns} or image processing techniques \cite{liu_2017}.

%\newpage

\section{Tactile Information Captured by the Representation}
\label{sec:properties}

\begin{table*}[t]
	\centering
	\caption{Information on taxels' location, measurements and distribution encoded by the representations}
	\label{tb:local_prop}
	\setlength\extrarowheight{2pt}
	%\resizebox{\textwidth}{!}{
	\begin{tabular}{c|ccccccl|}
		\cline{2-8}
		\multicolumn{1}{l|}{}                                                                                         & \multicolumn{7}{c|}{\textbf{Data Structure}}                                                                                                                                                                                                                                                                                                                                                                                                                                                                \\ \hline
		\multicolumn{1}{|c|}{\multirow{2}{*}{\textbf{\begin{tabular}[c]{@{}c@{}}Taxels'\\ Information\end{tabular}}}} & \multicolumn{1}{c|}{\multirow{2}{*}{\textit{Vector}}} & \multicolumn{1}{c|}{\multirow{2}{*}{\textit{Matrix}}} & \multicolumn{1}{c|}{\multirow{2}{*}{\textit{\begin{tabular}[c]{@{}c@{}}Point\\ Cloud\end{tabular}}}} & \multicolumn{1}{c|}{\multirow{2}{*}{\textit{Mesh}}} & \multicolumn{1}{c|}{\multirow{2}{*}{\textit{Image}}} & \multicolumn{2}{c|}{\textit{Map}}                                                                                                                                            \\ \cline{7-8} 
		\multicolumn{1}{|c|}{}                                                                                        & \multicolumn{1}{c|}{}                                & \multicolumn{1}{c|}{}                                 & \multicolumn{1}{c|}{}                                                                                & \multicolumn{1}{c|}{}                                & \multicolumn{1}{c|}{}                                & \multicolumn{1}{c|}{\textit{\begin{tabular}[c]{@{}c@{}}From \\ Raw\end{tabular}}} & \multicolumn{1}{c|}{ \textit{\begin{tabular}[c]{@{}c@{}}From\\ Calibration\end{tabular}} } \\ \hline
		\multicolumn{1}{|c|}{\textit{\begin{tabular}[c]{@{}c@{}}Raw\\ Measurements\end{tabular}}}              & \multicolumn{1}{c|}{$\CIRCLE$}                                & \multicolumn{1}{c|}{$\CIRCLE$}                                 & \multicolumn{1}{c|}{$\CIRCLE$}                                                                                & \multicolumn{1}{c|}{$\CIRCLE$}                                & \multicolumn{1}{c|}{$\CIRCLE^{*1}$}                                & \multicolumn{1}{c|}{$\Circle^{*2}$}                                                                    &              $\Circle^{*2}$                                                                     \\ \hline
		%
		%\multicolumn{1}{|c|}{\textit{\begin{tabular}[c]{@{}c@{}}Temporal\\ Information\end{tabular}}}              & \multicolumn{1}{c|}{$\CIRCLE$}                                & \multicolumn{1}{c|}{$\CIRCLE$}                                 & \multicolumn{1}{c|}{$\CIRCLE$}                                                                                & \multicolumn{1}{c|}{$\CIRCLE$}                                & \multicolumn{1}{c|}{$\CIRCLE^{*1}$}                                & \multicolumn{1}{c|}{$\Circle^{*2}$}                                                                    &              $\Circle^{*2}$                                                                     \\ \hline
		%
		\multicolumn{1}{|c|}{\textit{\begin{tabular}[c]{@{}c@{}} Locations\end{tabular}}}                   & \multicolumn{1}{c|}{}                                & \multicolumn{1}{c|}{}                                 & \multicolumn{1}{c|}{\CIRCLE}                                                                                & \multicolumn{1}{c|}{\CIRCLE}                                & \multicolumn{1}{c|}{$\Circle^{*3}$}                                & \multicolumn{1}{c|}{}                                                                    &                       $\Circle^{*2}$                                                            \\ \hline
		\multicolumn{1}{|c|}{\textit{\begin{tabular}[c]{@{}c@{}}Spatial \\ Relations\end{tabular}}}                   & \multicolumn{1}{c|}{}                                & \multicolumn{1}{c|}{}                                 & \multicolumn{1}{c|}{\CIRCLE}                                                                                & \multicolumn{1}{c|}{\CIRCLE}                                & \multicolumn{1}{c|}{$\CIRCLE^{*4}$}                                & \multicolumn{1}{c|}{}                                                                    &                           $\CIRCLE^{*4,5}$                                                        \\ \hline
		\multicolumn{1}{|c|}{\textit{\begin{tabular}[c]{@{}c@{}}Topological \\ Relations\end{tabular}}}               & \multicolumn{1}{c|}{}                                & \multicolumn{1}{c|}{}                                 & \multicolumn{1}{c|}{\CIRCLE}                                                                                & \multicolumn{1}{c|}{\CIRCLE}                                & \multicolumn{1}{c|}{$\CIRCLE^{*6}$}                                & \multicolumn{1}{c|}{$\CIRCLE^{*5,6}$}                                                                    &                                               $\CIRCLE^{*5,6}$                                    \\ \hline
	\end{tabular}
		%}
	\begin{tablenotes}
		\item[*] The table highlights which taxels' information is encoded by the various data structures. In this respect, \CIRCLE~ refers to a property directly encoded, while \Circle~ indicates the need for an additional data structure or computation to retrieve the property.  It can be seen that point clouds and meshes are equivalent from this point of view, no matter the hardware. \\ 
		$^1$ Approximated due to the interpolation;
		$^2$ Can be computed if the mapping is invertible;
		$^3$ Pixels need to be associated to coordinates in space. If the taxel distribution does not resemble a planar grid, only an approximation can be computed.
		$^4$ Not preserved when deformations are introduced or cuts are required; 
		$^5$ The mapping must be designed to preserve the property; 
		$^6$ Not preserved when cuts are required. 
	\end{tablenotes}
\end{table*}

After presenting the state-of-the-art methods for representing tactile data and discussing how hardware aspects influence their selection, this section addresses how representing data with a particular structure implicitly filters tactile information, thereby affecting which contact properties can be perceived.

Having introduced the generic structure of the hardware and the concept of taxel in Section 4, %\cref{sec:hardware}, 
it becomes clear that, task-level information of interest such as contact forces, positions, shape or dynamics can all be computed from the knowledge of taxels' information, including raw data (and their evolution over a time window of interest), location in space and distribution.
As shown in the following analysis, depending on the data structure used as the representation, it is not always possible to preserve all taxels' information.
This strongly impacts the whole tactile perception pipeline shown in \cref{img:overview} and directly affects the task execution. Since the representation should be functional for the task, understanding which taxels' information is \textit{encoded} becomes extremely relevant. 

This section analyzes the types of information that can be captured by different data structures, including raw measurements, temporal dynamics, contact locations, and spatial distribution, and discusses their implications for tactile perception.
The outcome of this analysis is reported in \cref{tb:local_prop}.
Full circles indicate structures that directly encode the information, while empty circles refer to the possibility of retrieving the information with the support of additional data structures or computation.

\subsection{Raw Measurements}
This refers to whether the data structure explicitly preserves the raw measurement associated with each individual taxel.
It allows for computing information related to the magnitude of forces or pressures acting on the compliant layer, as well as tracking its dynamics over time.
This is encoded by all structures. 
Vectors and matrices are containers filled with raw measurements. 
This information can also be associated with elements in the point cloud \cite{albini_2021_2} or with nodes in meshes \cite{fan_2022,albini_2021}. 
For images, the value of the single pixel is related to the raw measurement. When the taxels' spatial distribution does not match a grid pattern and resampling is needed (see \cref{img:layout2representation}), the value encoded by the pixel approximates the original measurement. 
In the case of maps, it is possible to preserve the original measurements as long as the mapping between the 3D and planar domain is invertible \cite{denei_2014,sutanto_2019}, otherwise they can be approximated \cite{pugach_2016}.

This discussion applies when the time evolution of raw measurements is considered. In particular, two main types of temporal information can be encoded:
\begin{itemize}
	\item \textbf{Fixed-interval sampling}: Measurements are collected at a fixed sampling rate.
	\item \textbf{Event-based sampling}: Data is collected only when specific conditions are met. This can occur at the hardware level \cite{bartolozzi_2017,liu_2022}, or be informed by the task (e.g., capturing data at steady-state in different grasp conditions, as in \cite{shorthose_2023}).
\end{itemize}
In this respect, vectors and matrices can store measurements corresponding to different time instants (see \cref{img:matrix}). Images can be extended by adding temporal channels (analogous to RGB channels in visual data) to encode the sequence of taxel responses \cite{chaoyi_2023}.
To the best of the authors’ knowledge, point clouds have not yet been widely adopted for temporal encoding, although this is technically feasible.
As with image-based representations, multiple channels can be associated with each point. These multi-channel point clouds could then serve as a basis for constructing meshes that embed time-dependent information. The same principle applies to maps, which are often implemented as planar point clouds or graph structures.

\subsection{Locations}
\label{subsec:properties_locations}
A representation encoding the location of taxels in the form of Cartesian coordinates allows for computing task-level information related to contact locations or centroids.
A requirement for this is having spatially calibrated taxels. Point clouds and meshes belong to this category.

Taxels in tactile images are indexed with rows and columns and their Cartesian coordinates are not directly encoded in the structure. 
However, they can still be retrieved by associating coordinates to the pixels.
When the taxels' spatial distribution does not follow a grid-like pattern, the original locations are lost due to the resampling operation shown in the right part of \cref{img:layout2representation}.
However, they can still be approximated by back-projecting the pixel indexes to the original 3D space (assuming the flattening operation to be invertible). 

Vectors, matrices and maps cannot encode the locations as they are just a way to reorganize the taxels' raw measurements. 
However, if the map is created from a spatially calibrated sensor, such as in \cite{sutanto_2019,denei_2014}, it is possible retrieve the taxels' location if the mapping operation is invertible.

\subsection{Spatial Distribution} 
A representation can capture two properties related to the taxels' distribution - spatial and topological relations. 
A structure encoding the \textit{spatial relations} maintains the relative distances among the taxels
\footnote{It is worth noting that, unlike taxels' locations, which specify their exact Cartesian coordinates, spatial relations capture only the relative distances. As a result, the taxel distribution can be uniformly scaled without affecting these relations.}
Therefore, when a force is applied, taxels can provide discrete information on how the contact is distributed, thus allowing for reconstructing its shape and the dynamics over time and space.
Encoding the spatial relations directly implies the possibility of reconstructing the contact shape.
\textit{Topological relations}, on the other hand, are related to the adjacencies among the single taxels or a group, regardless of their relative distances. 
 
\begin{figure}[t!]
	\centering
	\subfigure[]{\includegraphics[width=0.48\textwidth]{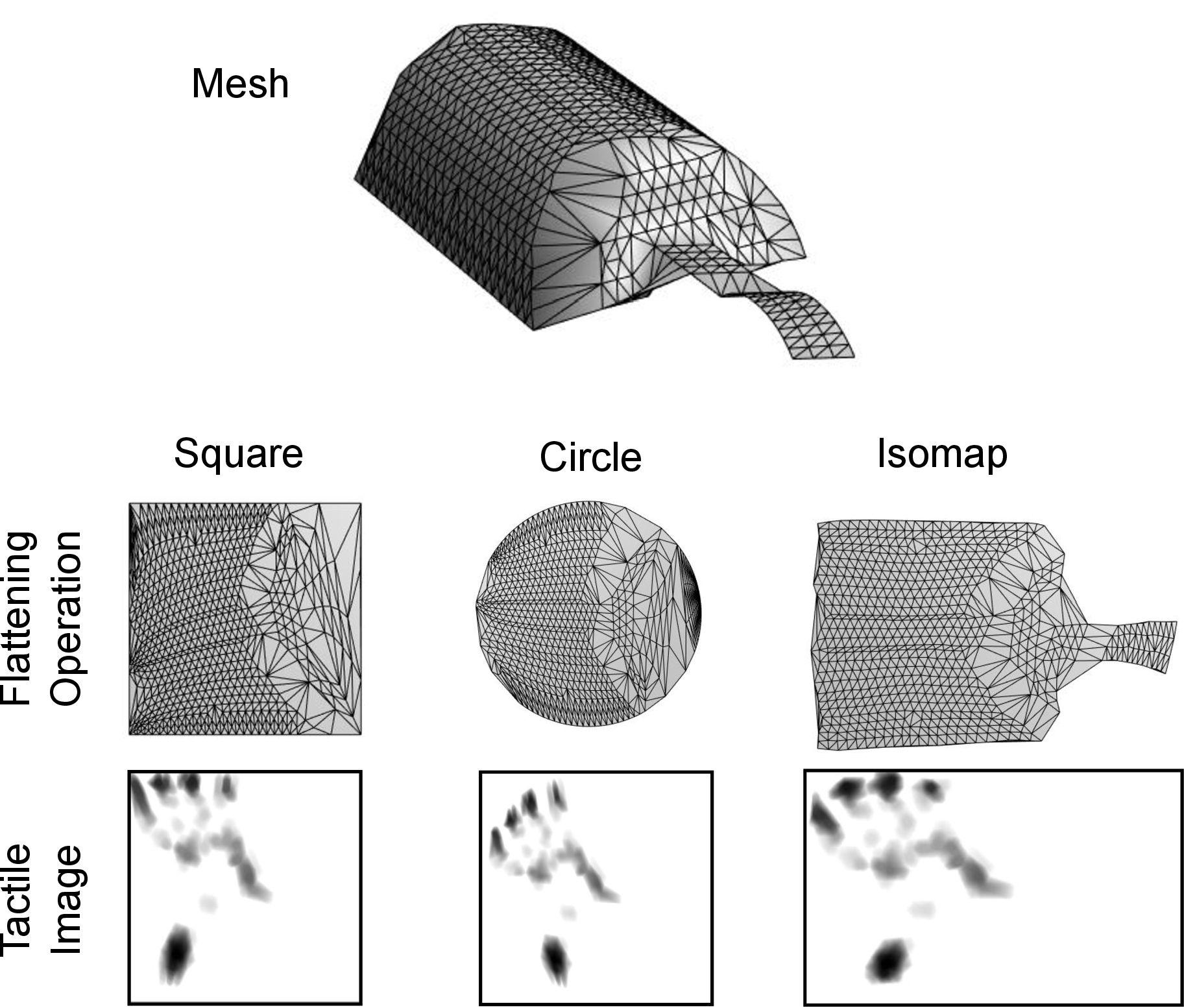} \label{img:distorsion}} \quad
	\subfigure[]{\includegraphics[width=0.48\textwidth]{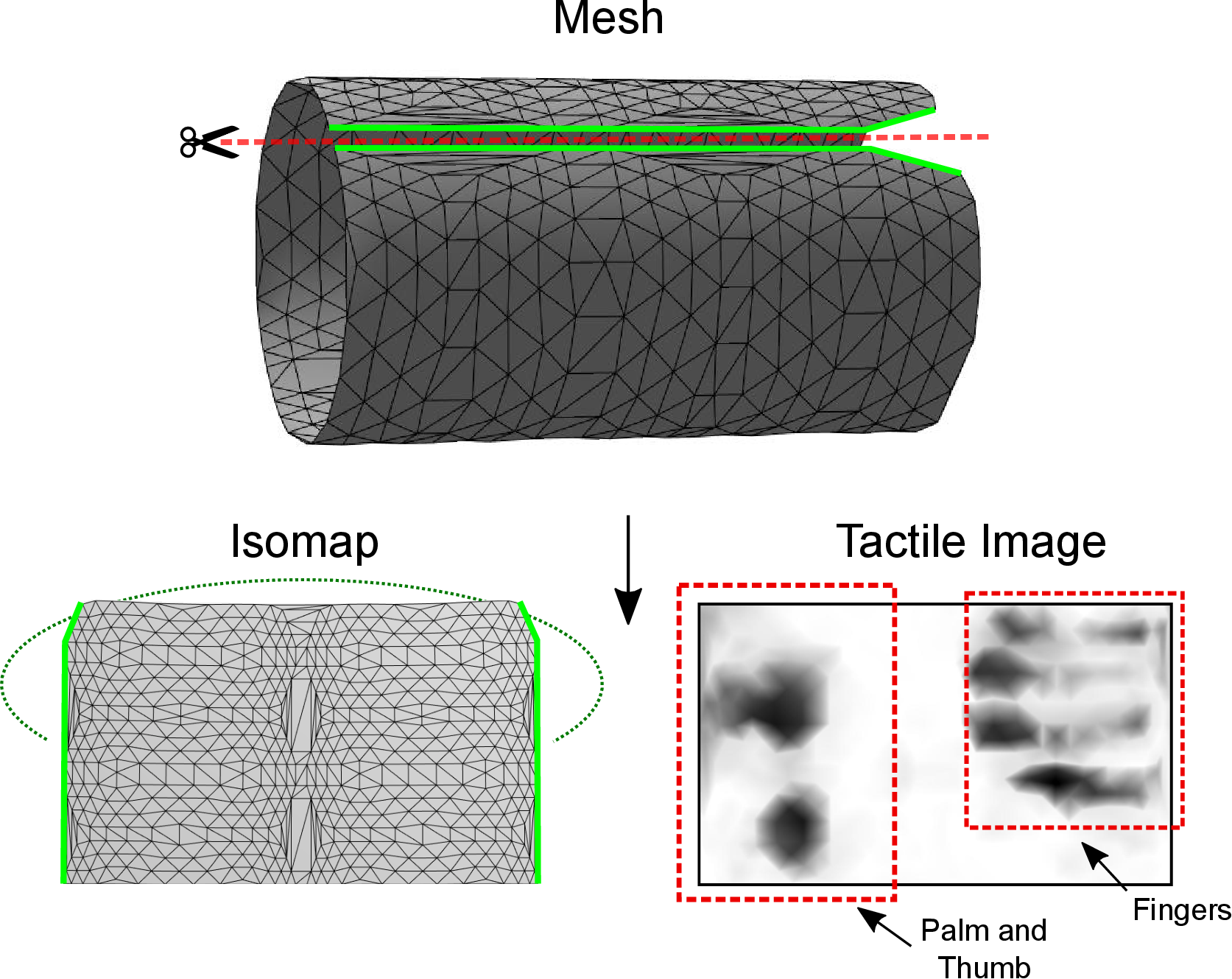} \label{img:cuts}}
	\caption{Issues occurring when creating tactile images starting from complex taxels distributions. (a) Three different methods to flatten a non-planar distribution of taxels. As shown, the geometry of the skin is distorted. This is also reflected in the corresponding tactile images of the hand grasping the surface. (b) The effect of cuts to open non-planar surfaces. The geometry at the boundary of the cut cannot be preserved in the 2D space, thus splitting the resulting contact shape. 
	}
	\label{img:singularities}
\end{figure}

\subsubsection*{Spatial Relations}

Point clouds and meshes encode this property since it is directly related to the spatial calibration providing a 1-to-1 correspondence between the taxels and points. 
Images encode this property if generated from taxels having a flat distribution - the relative position and distances among pixels correspond to a resampled and scaled version of the geometry of the sensor.
However, for non-planar spatial distributions, 
%\cref{sec:build} 
the flattening operation described in Section 5 %\cref{sec:build} 
may introduce distortions whose strength depends on the flattening algorithm and on the complexity of the original taxels' spatial distribution. 
An example is shown in \cref{img:distorsion}. The taxels' placement in 3D (represented as a mesh) corresponds to a real-world example where 678 tactile sensors have been integrated into a Baxter robot \cite{albini_2020}. The image shows three different flattening operations and the resulting images, obtained when a human hand is grasping the robot.  As shown, the fingers of the hand are distorted depending on the way the mesh is flattened.
Although distortions may be mitigated with the choice of a suitable flattening algorithm, they are not generally avoidable \cite{desbrun_2002}. Therefore, as a side effect, the relative distances are not correctly encoded in some areas of the image. \\
An additional challenge is introduced when cuts are required to unfold the surface. An example with a cylinder is shown in \cref{img:cuts}. Taxels close together in 3D are on the opposite side in the corresponding 2D representation. 
As shown in \cref{img:cuts}, this affects the contact shape reconstructed by the tactile image - the hand grasping the cylinder is split into two halves, and the representation does not correctly encode the original contact distribution at the boundaries of the cut. 

Maps can be designed to preserve the geometry only if they are built from the knowledge of the original taxels' distribution. For example, 
\cite{denei_2014} leverages a transformation maintaining the spatial relations among taxels. However, being this a flattening operation, the same issue previously described for tactile images occurs.  

Vector and matrix representations lack information about the taxel relations, and as such cannot encode this property. 

\subsubsection*{Topological Relations}

This refers to the possibility of encoding the adjacencies among the taxels without considering their distances. 
This information is preserved by the point cloud and explicitly encoded through the edges in meshes.
For images, the flattening of a curved surface maintains the adjacencies (even if the distances may be lost due to distortion) when open manifolds are considered (see \cref{img:distorsion}). 
However, when cuts are required (as in \cref{img:cuts}), the topology is not preserved for the taxels lying at the edge of the cut.

Maps can be designed to preserve the topology as shown in \cite{hoffmann_2018,brock_2009,pugach_2016,pugach_2015}. The training procedure can group taxels into clusters correlating measurements of sensors stimulated at the same time. 
For example in \cite{hoffmann_2018,brock_2009} taxels belonging to different robot body parts are semantically clustered. The work in \cite{pugach_2016,pugach_2015} proposes a grid-like \gls{som} resembling a tactile image, preserving the topological configuration of the taxels.

\subsection{Discussion}

The analysis reported in this Section provides a number of insights for selecting a representation depending on the taxel-level information that needs to be encoded. From \cref{tb:local_prop} it can be seen that not all the structures introduced in Section 5 %\cref{sec:build} 
encode all information related to taxels. From the point of view of task-level requirements, this is still acceptable; often only a subset of the information may be required. 	%
For instance, tactile-based gesture recognition can be achieved from the knowledge of taxels' measurements alone  \cite{tawill_2014} or topology \cite{brock_2009}.
On the contrary, tactile-based control also requires information on taxels' locations \cite{jain_2013,albini_2021_2,killpack_2016}. 
Similarly, for object recognition, the whole contact shape must be processed and the precise taxels' locations in space are not relevant in general \cite{schneider_2009,liu_2012,luo_2015,gandarias_2018,luo_2016,liu_2012_1}.

For tasks requiring contact information retrievable from the taxels' measurements only (or from their evolution over time), vectors and matrices are the most immediate structures to apply. They can be selected depending on the type of processing required by the task. For example, feature extraction based on descriptors is commonly performed on vectors of time series \cite{khan_2016, wang_2022,fishel_2012, kaboli_2015}, while matrices are worth considering to apply \gls{cnns} \cite{shorthose_2023,gao_2015}.

When topological information is required and accurate spatial calibration is hard to obtain (see Appendix), maps are a valid option. 
Their construction usually involves a training phase, which is needed to build topological relations among the taxels. In this respect, data required by the training procedure must be properly collected. 
For instance, the map in \cite{kuniyoshi_2004} is trained to preserve the topology when taxels are stimulated together as the effect of the same contact event. However, if more than one contact simultaneously occurs on different (and far) body locations, contacts are represented close in the map while the real contact locations are far from each other.

Point clouds and meshes can encode all the information related to the taxels' locations, measurements and distribution, regardless of the hardware.
The same applies to images when the taxels are arranged into a planar grid. 
Otherwise, as previously discussed, partial loss of information may occur when mapping from the 3D to the 2D domain (see \cref{img:singularities}).
It is important to note that distortions or cuts related to the flattening operation are not always an issue. 
This mainly depends on the high-level computation that needs to be applied. 
As an example, a classification task based on \gls{ml} would be robust to distortions because the contact shape is always mapped in a repeatable manner into the image \cite{albini_2020}.
On the contrary, in control tasks, distortions can affect the performance of the robot controller and cuts lead to discontinuities. Therefore, in this context, the choice of an appropriate mapping algorithm becomes more important. 

It is also important to remark that tasks requiring the processing of the contact shape can be supported only by representations encoding spatial relations. Therefore, point clouds, meshes, images or maps are the only ones worth considering.

\section{Data Representation for Processing, Inference and Action Operations}

\label{sec:tasks}

\begin{table*}[]
	\centering
	\caption{Operations exploiting tactile feedback in relation to the corresponding data representation}
	\label{tb:repr_tasks}
	\setlength\extrarowheight{2pt}
	%	\resizebox{\textwidth}{!}{
		\begin{tabular}{cc|cccccc|}
			\cline{3-8}
			&                                                                                      & \multicolumn{6}{c|}{\textbf{Data Structure}}                                                                                                                                                                                                                   \\ \cline{3-8} 
			\multicolumn{2}{c|}{\textbf{}}                                                                                                                    & \multicolumn{1}{c|}{\textit{Vector}} & \multicolumn{1}{c|}{\textit{Matrix}} & \multicolumn{1}{c|}{\textit{\begin{tabular}[c]{@{}c@{}}Point\\ Cloud\end{tabular}}} & \multicolumn{1}{c|}{\textit{Mesh}} & \multicolumn{1}{c|}{\textit{Image}} & \textit{Map}          \\ \hline
			\multicolumn{1}{|c|}{\multirow{4}{*}{\textbf{Processing}}} & \textit{\begin{tabular}[c]{@{}c@{}}Computation of Contact\\ Parameters\end{tabular}} & \multicolumn{1}{c|}{}               & \multicolumn{1}{c|}{}                & \multicolumn{1}{c|}{$\CIRCLE$}                                                      & \multicolumn{1}{c|}{$\CIRCLE$}     & \multicolumn{1}{c|}{$\Circle$}      & $\Circle$             \\ \cline{2-8} 
			\multicolumn{1}{|c|}{}                                     & \textit{\begin{tabular}[c]{@{}c@{}}Force Field\\ Reconstruction\end{tabular}}        & \multicolumn{1}{l|}{}               & \multicolumn{1}{l|}{}                & \multicolumn{1}{l|}{}                                                               & \multicolumn{1}{c|}{$\Circle$}     & \multicolumn{1}{c|}{$\CIRCLE$}      & \multicolumn{1}{l|}{} \\ \cline{2-8} 
			\multicolumn{1}{|c|}{}                                     & \textit{Shape Reconstruction}                                                       & \multicolumn{1}{c|}{}               & \multicolumn{1}{c|}{}                & \multicolumn{1}{c|}{$\CIRCLE$}                                                      & \multicolumn{1}{c|}{}              & \multicolumn{1}{c|}{$\Circle$}      &                       \\ \cline{2-8} 
			\multicolumn{1}{|c|}{}                                     & \textit{Feature Extraction}                                                         & \multicolumn{1}{c|}{$\CIRCLE$}      & \multicolumn{1}{c|}{}                & \multicolumn{1}{c|}{}                                                               & \multicolumn{1}{c|}{$\Circle$}     & \multicolumn{1}{c|}{$\CIRCLE$}      &                       \\ \hline
			\multicolumn{1}{|c|}{\multirow{6}{*}{\textbf{Inference}}}  & \textit{Object Classification}                                                       & \multicolumn{1}{c|}{}               & \multicolumn{1}{c|}{$\Circle$}       & \multicolumn{1}{c|}{$\Circle$}                                                               & \multicolumn{1}{c|}{}     & \multicolumn{1}{c|}{$\CIRCLE$}      &                       \\ \cline{2-8} 
			\multicolumn{1}{|c|}{}                                     & \textit{\begin{tabular}[c]{@{}c@{}}Textures and Materials\\ Recognition\end{tabular}} & \multicolumn{1}{c|}{$\CIRCLE$}      & \multicolumn{1}{c|}{$\Circle$}       & \multicolumn{1}{c|}{}                                                               & \multicolumn{1}{c|}{}              & \multicolumn{1}{c|}{$\CIRCLE$}      &                       \\ \cline{2-8} 
			\multicolumn{1}{|c|}{}                                     & \textit{Gesture Recognition}                                                         & \multicolumn{1}{c|}{$\CIRCLE$}      & \multicolumn{1}{c|}{}                & \multicolumn{1}{c|}{}                                                               & \multicolumn{1}{c|}{}              & \multicolumn{1}{c|}{$\CIRCLE$}      & $\Circle$             \\ \cline{2-8} 
			\multicolumn{1}{|c|}{}                                     & \textit{Slip Detection}                                                              & \multicolumn{1}{c|}{$\CIRCLE$}      & \multicolumn{1}{c|}{}                & \multicolumn{1}{c|}{}                                                               & \multicolumn{1}{c|}{}              & \multicolumn{1}{c|}{$\CIRCLE$}      &                       \\ \cline{2-8} 
			\multicolumn{1}{|c|}{}                                     & \textit{\begin{tabular}[c]{@{}c@{}}Geometric Features\\ Detection\end{tabular}}      & \multicolumn{1}{c|}{$\Circle$}      & \multicolumn{1}{l|}{}                & \multicolumn{1}{l|}{}                                                               & \multicolumn{1}{c|}{$\Circle$}     & \multicolumn{1}{c|}{$\CIRCLE$}      & $\Circle$             \\ \cline{2-8} 
			\multicolumn{1}{|c|}{}                                     & \textit{Pose Estimation}                                                             & \multicolumn{1}{c|}{}               & \multicolumn{1}{c|}{}                & \multicolumn{1}{c|}{$\CIRCLE$}                                                      & \multicolumn{1}{c|}{}              & \multicolumn{1}{c|}{$\CIRCLE$}      &                       \\ \hline
			\multicolumn{1}{|c|}{\multirow{4}{*}{\textbf{Action}}}     & \textit{\begin{tabular}[c]{@{}c@{}}Object Manipulation\\ and Grasping\end{tabular}}  & \multicolumn{1}{c|}{}               & \multicolumn{1}{c|}{}                & \multicolumn{1}{c|}{$\Circle$}                                                      & \multicolumn{1}{c|}{$\Circle$}     & \multicolumn{1}{c|}{$\CIRCLE$}      &                       \\ \cline{2-8} 
			\multicolumn{1}{|c|}{}                                     & \textit{\begin{tabular}[c]{@{}c@{}}Tactile-Driven \\ Exploration\end{tabular}}       & \multicolumn{1}{c|}{$\CIRCLE$}      & \multicolumn{1}{c|}{$\CIRCLE$}       & \multicolumn{1}{c|}{}                                                               & \multicolumn{1}{c|}{}              & \multicolumn{1}{c|}{}               &                       \\ \cline{2-8} 
			\multicolumn{1}{|c|}{}                                     & \textit{Compliant Control}                                                           & \multicolumn{1}{c|}{}               & \multicolumn{1}{c|}{}                & \multicolumn{1}{c|}{$\CIRCLE$}                                                      & \multicolumn{1}{c|}{}              & \multicolumn{1}{c|}{}               & $\Circle$             \\ \cline{2-8} 
			\multicolumn{1}{|c|}{}                                     & \textit{Whole Arm Control}                                                           & \multicolumn{1}{l|}{}               & \multicolumn{1}{l|}{}                & \multicolumn{1}{c|}{$\CIRCLE$}                                                      & \multicolumn{1}{l|}{}              & \multicolumn{1}{l|}{}               & $\Circle$             \\ \hline
		\end{tabular}
		%	}
	\begin{tablenotes}
		\item[*] The table highlights links between operations and representations to provide an immediate understanding of the type of data structure suitable for a given operation. Filled circles represent data structures used by the majority of works presented in the literature. Empty circles indicate alternatives considered by a significantly lower number of works.  
	\end{tablenotes}
\end{table*}

This section aims to analyse how Data Representation supports high-level operations related to Processing, Inference, and Action, as illustrated in \cref{img:overview}.
	\cref{tb:repr_tasks} includes a list of applications that have already been analysed in previous survey papers, such as \cite{luo_2017, wei_2018, qiang_2020, hu_2023}. This section reviews the same application domains, categorized into Processing, Inference, and Action stages as shown in \cref{img:overview}, but from the specific viewpoint of Data Representation.
	The aim is to examine how different structures support or constrain operations that leverage tactile sensing.
A discussion follows at the end of this section elaborating on how the representation can be selected depending on the required operation.

\subsection{Processing}

The operations described in the following apply processing on top of the data structures needed to compute contact parameters, to perform filtering, feature extraction, or to merge multi-contact information.

\subsubsection*{Computation of Contact Parameters}

Structured data is generally processed to compute contact parameters of interest for the task, such as the positions of multiple contacts with respect to a desired reference frame and the relative forces. % 
Contact locations on the robot surface are usually found by computing the contact centroid, i.e., the geometric centroid among the positions of taxels involved in the contact (possibly weighted by their measurements). This processing usually involves a previous step of thresholding to remove the baseline of the sensor output, and a further step of clustering to identify connected regions of taxels involved in the contact. Contact centroids (and their relative normals) can then be computed for each set of connected taxels. The equivalent force associated with the centroid can then be computed by averaging the measurements of the taxels belonging to the cluster (provided that the taxels are characterized or force-calibrated \cite{kangro_2017}). This operation can be performed with any representation encoding the taxels' location (see Section 6) % \cref{sec:properties})
, i.e., point clouds \cite{albini_2021_2,jain_2013,wosch_2002,schmidt_2006}, meshes \cite{quan_2023,park_2024}. Images and map structures can also be used, provided that the conditions described in Section 6.2 are met. % \cref{subsec:properties_locations} are met.

\subsubsection*{Force Distribution Reconstruction}

As explained in Section 4, %\cref{sec:hardware}, 
the output of a tactile sensor is a function of the deformation applied, however, its physical meaning depends on the specific transduction technology. Therefore, it may be of interest to retrieve the corresponding real forces or pressures from the raw sensor measurements. Although a force/pressure calibration of individual taxels is doable \cite{kangro_2017}, other methods focus on reconstructing the exact force distribution \cite{sato_2010, tawill_2011, muscari_2013, seminara_2015, guo_2016, ma_2019, wasko_2019, han_2025, zhang_2025}.  

Even though these works differ in terms of methodology, all of them are based on a geometric resampling of the sensing surface. Forces are computed at each sampling point by applying a contact model based on marker displacements \cite{sato_2010, guo_2016, ma_2019} or by explicitly modeling the physics of the sensor \cite{tawill_2011, muscari_2013, seminara_2015, wasko_2019}.
The resampling is commonly performed using a grid, thus leading to structures equivalent to images. However, \cite{tawill_2011} also demonstrates the possibility of resampling using a mesh structure.   

\subsubsection*{Shape Reconstruction}
% Shape reconstruction
Tactile data acquired from multiple interactions can be processed to reconstruct the overall shape of objects or the environment.
In this respect, two approaches have been proposed. 
The first aims to obtain a discrete representation of the object - by increasing the number of contacts, a set of small point clouds (representing the local deformation of the surface) is stored and merged to generate a larger point cloud encoding the whole shape of the object \cite{parsons_2022,falco_2019,sommer_2014,sommer_2016,meier_2011, kaboli_2019,murali_2023}. Alternatively, tactile images collected at each contact can also be used, but they need to be processed and converted into a small point cloud as shown in \cite{bauza_2019,zhang_2025}. 

The second approach is based on further processing of the resulting point cloud to obtain a continuous representation in the form of a surface. 
In this respect, \gls{gpis} \cite{driess_2017,chaoyi_2023,suresh_2021,ottenhaus_2018} and \gls{sdf} \cite{comi_2023} can be built on top of point clouds. An alternative technique is presented in \cite{ottenhaus_2016}, which, compared to \gls{gpis}, better reconstructs the shape of sharp geometric features of the target object (e.g., edges).

\subsubsection*{Features Extraction}

This type of processing aims to highlight and extract features of interest from tactile data. 
When time-based features are considered, time series of tactile data are arranged as vectors to analyze the evolution of the sensor output over time. Vectors are then processed to compute descriptors, which generally support further steps of inference based on \gls{ml}. 
Descriptors usually consist of statistical features computed over the time windows of interest, such as the average, maximum and minimum force, number of taxels involved, variations in the signal computed as first or second order derivatives of the time series.
Different descriptors can be computed for gesture recognition \cite{tawill_2014}, texture classification \cite{kaboli_2015,kaboli_2018}, or slip detection \cite{su_2015,choi_2005}.

A different type of processing has the goal of highlighting or extracting geometric features.
Typical operations involve applying filters to highlight features in the contact shape, such as edges, by processing tactile images \cite{berger_1991,chen_1995,gibbons_2023,pezzementi_2011} or meshes \cite{albini_2021,hongyu_2024}. 
Beyond the processing of the contact shape, the work in \cite{falco_2019} (related to object recognition) first reconstructs the shape of objects by merging multiple contact instances, then applies a feature extraction technique to compute geometric descriptors on the resulting point cloud.

\subsection{Inference}

It refers to operations aimed at performing classification, pose estimation, or, more generally, to algorithms that extract meaningful information to inform decisions on the next robot action. Some of these methods are grounded on a pre-processing of the data representation (see the previous Section). Others, mainly based on \gls{dl} solutions, integrate processing and inference stages within the same operation, thus directly working on the data representation \footnote{Minor processing operations can still be applied, such as baseline removal or thresholding.}. Examples include \gls{cnns}, where the same architecture performs feature extraction and classification.

\subsubsection*{Object Classification}
It consists of recognising objects by analysing properties that can be captured with the sense of touch, such as their local shape or mechanical properties. % that may not be directly visible. 
Object recognition is a well studied area in computer vision and has been expedited thanks to the progress in \gls{dl} for image classification \cite{voulodimos_2018}. 
As a consequence, it has become natural to represent tactile data in the form of images, encoding the contact shape, to leverage the use of \gls{cnns} methods \cite{schneider_2009,pezzementi_2011,liu_2012,luo_2015,gandarias_2018,luo_2016,bhattacharjee_2012,liu_2012_1,zhang_2025}.

A different approach consists in using representations based on matrices where raw taxels' measurements are reorganised. 
In \cite{shorthose_2023} vectors of tactile raw measurements acquired through multiple interactions are stacked. This data arrangement %enables the integration of multi-contact information into a grid structure, which 
can be effectively processed using \gls{cnns} to establish correlations among data acquired from various contacts. 
A similar approach was employed in \cite{gao_2015} to encode temporal data. Each column in the matrix
represents data at different instants in time. In this way, the convolution along the columns enable the extraction of temporally dependent features.

Although images are the most commonly used data representation for object recognition tasks \cite{liu_2017}, more recent approaches have begun investigating alternative solutions based on point clouds or meshes. %Similarly to images, the idea is to leverage existing algorithms already developed for different applications. 
For instance, multiple point clouds, each representing the taxels' location in space at a specific contact, can be combined to encode the whole object shape in the form of a point cloud \cite{zhang_2016,falco_2019,parsons_2022}, which can then be classified by computing geometric descriptors or using \gls{dl} techniques.
Moreover, the recent introduction of \gls{gcnns} has enabled classification tasks on mesh data structures \cite{wu_2021}. 

\subsubsection*{Textures and Materials Recognition}
This is generally based on two different approaches. 
The first analyzes the temporal or frequency information of the tactile signal resulting from the interaction with the surface. Typically, it requires active robot motions, such as sliding, to collect tactile samples. %These approaches usually employ sensors capturing high-frequency information. 
The rationale behind the approach is that the interaction with surfaces having different textures or materials produce distinct temporal or frequency patterns in the tactile signal. 
Raw taxel measurements are considered over a time window of interest and stored in vectors to be processed. Features are then extracted from vector or matrix structures and used to train \gls{ml} classifiers \cite{kaboli_2016,khan_2016, fishel_2012, kerr_2018}. \gls{cnns} can also be used for feature extraction, but this requires reshaping time series into a matrix as shown in \cite{baishya_2016,wang_2022}.

The second approach focuses on the classification of geometric features. These techniques process the contact shape to encode the surface texture. 
Classification is then performed on data represented as images, usually on top of a processing stage aimed at extracting features \cite{adelson_2013,bauml_2019,drimus_2011}, or by means of \gls{dl} solutions \cite{yuan_2018,taunyazov_2019}. 

\subsubsection*{Gesture Recognition}
This is particularly relevant in \gls{phri} and teaching by demonstration, where it may be more intuitive for the operator to communicate with the robot through physical interactions rather than using visual gestures.
Similarly to the previous case, there is a clear distinction between methods processing the dynamics of the contact and those that focus on its shape. 
Indeed, the first is based on the assumption that gestures can be recognized by analysing their temporal dynamics or the number of sensors involved. 
The general approach is similar to most of the techniques proposed in the literature \cite{tawill_2014}. In this case, time series of tactile data represented as vectors are processed by computing a set of features or descriptors which are then classified using supervised \gls{ml} algorithms \cite{naya_1999,stiehl_2005,huisman_2013}. 

Conversely, approaches based on contact shape processing assume different gestures lead to distinct pressure patterns collected by the tactile sensor. 
These methods rely on processing or classifying tactile images \cite{albini_2020,salvato_2021,cirillo_2017,hughes_2018,gandarias_2018,choi_2022,tawil_2012}. While most of these methods use images to model static information, \cite{choi_2022} demonstrated the possibility of stacking multiple tactile images as different channels, effectively encoding temporal information within a single 3D image structure. 

Beyond vectors or images, \cite{brock_2009} explored a map-based representation for the classification of haptic interactions. 
The map is learned by physically interacting with the tactile sensor and clusters the taxels into semantic groups (e.g. arm, torso, etc.) mapped into a plane. 

\subsubsection*{Slip Detection}
Tactile sensors have been extensively used to detect slips while grasping or manipulating objects and to provide feedback for better regulation of applied contact forces \cite{wei_2018}.
One possibility to detect slips consists of analysing the temporal dynamics of the tactile signal and looking for variations or peaks. 
These methods typically process vectors of time series of data either in time or frequency domain \cite{choi_2005,schoepfer_2010,kaboli_2016_1,meier_2016,fernandez_2014,su_2015,veiga_2018,barrett_2016}. Such approaches rely on high-frequency response tactile sensors, where the slip detection can be achieved using either model-based \cite{fernandez_2014,su_2015,choi_2005} or \gls{ml}  \cite{schoepfer_2010,barrett_2016,meier_2016,veiga_2018} techniques. 

A different class of methods identifies slips by analyzing variations in either the contact shape or the positions of taxels embedded into the soft layer. To date, we found images to be the only representation used by these approaches \cite{james_2018,ho_2012,watanabe_2007,ito_2011,yuan_2015,sui_2021}.

It is also worth mentioning a recent study~\cite{elijah_2025} that investigates how tactile data representation impacts slip detection, comparing vector-based structures with various types of tactile images. The results show that representations capturing spatial information lead to significantly higher slip detection accuracy, albeit at the cost of increased computational time.

\subsubsection*{Geometric Features Detection}

Identifying local geometric features of objects such as edges, contours or ridges, is essential for designing algorithms that enable robots to explore unknown objects, for example, by following contours. 
Early works addressing this problem used images to encode the contact shape, showing that classical image processing algorithms are suitable for detecting edges \cite{muthukrishnan_1987,berger_1991,chen_1995}. Nowadays, images are still the most used representation for this task. Recent works exploit them as a main representation to find features using model-based \cite{qiang_2013, tri_2010,lepora_2017,she_2021,gibbons_2023} or \gls{dl} approaches \cite{lepora_2019}.

Beyond images, alternative representations based on meshes and vectors have been explored. 
\cite{albini_2021} has shown the effectiveness of utilizing mesh representations for detecting contact features for sensors having a non-planar arrangement of the taxels. 
Furthermore, \cite{martinez_2013,martinez_2017} demonstrated the possibility of detecting the presence of edges directly from raw tactile data. These methods are based on Bayesian frameworks integrated with control strategies to actively discriminate features of interest. 

A work leveraging a map is presented in \cite{sutanto_2019} where a contour following task is performed on top of a representation mapping the non-planar taxels' distribution into a 2D latent space.

\subsubsection*{Pose Estimation}
 
Despite the variety of proposed approaches dealing with this task, all of them are grounded on two representations - images and point clouds.
Examples exploiting tactile images can be found in \cite{bauza_2020,gibbons_2023,sodhi_2021,li_2014,chebotar_2014,sodhi_2022,lepora_2021,lloyd_2024}. Data is processed using both model-based or \gls{dl} techniques in order to extract the relative position of the object to perform localization with respect to a known reference frame. 
A point cloud-based representation is exploited by several works \cite{bauza_2019,corcoran_2010,bimbo_2016,vezzani_2017,koval_2013,bimbo_2015,murali_2023}, and is mostly used to feed particle filter algorithms \cite{bimbo_2015,koval_2013,vezzani_2017,corcoran_2010}. 
A different processing method is presented in \cite{bimbo_2016} where the point cloud representation is used to match contact points with features describing the local surfaces in contact with the sensor. 

\subsection{Action}

This section discusses applications where tactile sensing is used as feedback to support the execution of control tasks related to object manipulation and interaction with humans or the environment. %\\

\subsubsection*{Object Manipulation and Grasping}

Tactile sensing has been shown to play a crucial role when applied to in-hand manipulation tasks. 
In this respect, tasks related to grasping \cite{si_2022,kolamuri_2021,zhao_2025}, object manipulation \cite{cherrier_2016,lambeta_2020,cherrier_2017,she_2021} or peg insertion \cite{gibbons_2023,kim_2022} are generally supported by images, as the encoded contact shape is extracted and processed through Processing and Inference operations to support decision-making and command a suitable control action.
Beyond tactile images, other approaches leverage the information related to the taxel positions encoded as point clouds to compute the contact locations.
Examples of works exploiting this representation can be found in the context of grasping \cite{miao_2014} or insertion tasks \cite{chaoyi_2023}. 

More recent approaches demonstrated the use of \gls{gcnns} in grasping tasks, where the taxels' distribution is represented using meshes encoding positions and adjacencies of taxels integrated into the dexterous hands \cite{funabashi_2022,yang_2023} or fingertips \cite{garcia_2019}. 
	As the mesh explicitly encodes both spatial and topological information, it allows the learning process to extract geodesic features over the tactile surface and to preserve the local neighbourhood relationships among taxels embedded in different fingers. 

\subsubsection*{Tactile-Driven Exploration}

It refers to robot actions whose motion is parameterized by the Inference step with the goal of exploring objects or surfaces. 
	These approaches do not generally require spatial information, as they are based on Bayesian exploration frameworks working on time series of raw data previously processed. 
	As discussed in Sections 5 and 6, vectors and matrices are particularly suitable for encoding temporal information over a specific window.
	Examples of their usage
can be found in the context of contour following \cite{martinez_2013,martinez_2017} supported by tactile data represented as vectors encoding time series. 
Matrix structures can be used as well as shown by \cite{scimeca_2020,scimeca_2020_1,scimeca_2018,fishel_2012}, where the control parameters are tuned on the basis of the Inference step, to perform the best action helping to discriminate among different object properties. 

\subsubsection*{Compliant Control} 

It refers to tasks dealing with the problem of interacting with robots using tactile feedback. 
Applications related to \gls{phri} usually exploit the sense of touch to make the robot compliant to the human input as shown in \cite{wosch_2002,schmidt_2006,frigola_2006,albini_2017}.
These methods use the information on (possibly) multiple contact locations to compute suitable motions in terms of directions and velocities, to follow the human-applied force. 
	Contact locations are computed from taxel coordinates in Cartesian space. Their measurements, representing the intensity of the contact, are then associated with each point. According to the discussion in Section 6, a point cloud is the most immediate representation in this context.

A map representation has been leveraged by \cite{pugach_2016} to implement a touch-based admittance control by transforming raw tactile data into a \gls{som}. The map in this case avoids the explicit modelling of the surface embedding the sensors, providing a lower dimensional implicit representation. The location of the tactile stimuli can be retrieved on the \gls{som} and used to command a proper robot motion.

\subsubsection*{Whole Arm Control}

This class of tasks addresses the problem of controlling a robot arm in a safe manner by exploiting tactile feedback. 
While the majority of the approaches previously discussed considers sensors mainly integrated into fingertips or hands, these works use tactile sensing distributed over large parts of the robot body.
These applications implement controllers using tactile feedback to regulate the contact forces applied on the environment either using model-based \cite{jain_2013,killpack_2016,albini_2021_2} or data-driven \cite{calandra_2015} approaches. 
Despite differences in terms of the implemented control scheme, all of them require knowledge of the locations and intensities of contact forces with respect to the robot base, that can be easily obtained from a point cloud representation of the taxels.

A different work simultaneously exploits two models - a mesh representing the 3D locations of the taxels and a map encoding a planar representation of the same mesh. The two are used to plan a hybrid position/force trajectories in a lower dimensional space. The trajectory is then back-projected to the mesh to be executed by the robot \cite{denei_2014}. 

\subsection{Discussion}

\begin{figure*}[t!]
	\centering
	\includegraphics[width=0.98\textwidth]{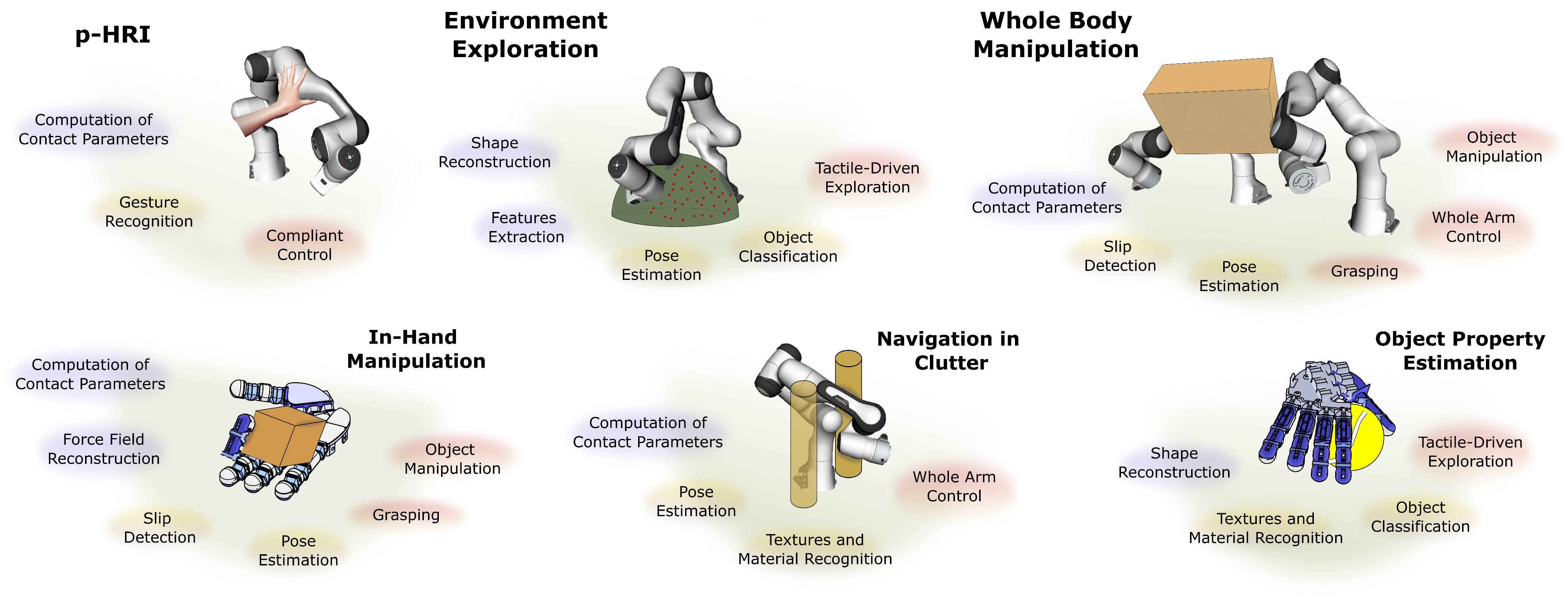}
	\caption{
		Examples of possible tasks exploiting tactile feedback. The whole task requires one or more operations related to Processing (blue), Inference (yellow) and Action (red). The different operations match the entries in \cref{tb:repr_tasks}.
	}
	\label{img:tasks}
\end{figure*}

\cref{tb:repr_tasks} shows an overview of what is discussed in this Section, showing which representation should be considered depending on the specific operations required by the task.
Filled circles indicate the data structures commonly used in the literature for the given operation, while empty ones represent alternatives considered by a significantly lower number of papers
\footnote{
	This categorization is qualitative.
	In most cases, one or two representations clearly dominate ($\sim$70\% of relevant works), 	while others are significantly less common ($\sim$30\%). The ratio may vary depending on the application.
}.

\subsubsection*{Application-Driven Selection}

From an application-driven perspective, some general guidelines can be identified.
Tactile images and point clouds stand out as the most commonly employed data structures across various tasks requiring information on the contact distribution. This is justified by the well-established history of data processing in computer vision spanning decades. Additionally, a wealth of processing algorithms is readily available to the robotics community through open-source libraries like OpenCV \cite{opencv_library} and PCL \cite{Rusu_ICRA2011_PCL}. The recent advancements in \gls{dl} for computer vision have given rise to \gls{dl}-networks that can be readily applied to images or point clouds out of the box \cite{8016501,qi2016pointnet}. 
Leveraging these data structures enables the direct application of existing algorithms to the tactile domain with minimal effort. Likewise, matrix-based representations, employed in object classification, are designed with the same objective of reusing \gls{cnns}. Therefore, when an application mirrors a computer vision problem or requires the use of well-established state-of-the-art methods developed for vision, tactile images, matrices, or point clouds should be taken into consideration.

Vectors of time series are particularly useful in tactile-based robotic tasks where the temporal dynamics of contact play a crucial role and the information on the contact position in the space is not relevant. 
When geometric features, contact locations or forces need to be extracted from non-planar or sparse sensor distributions, or data needs to be classified with \gls{gcnns}, meshes provide a representation where local adjacencies are explicitly encoded and ready to be processed.
This can bring advantages as shown by \cite{garcia_2019}, where the interplay meshes-\gls{gcnns} led to better performance with respect to the use of images-\gls{cnns}. From a computational perspective, \cite{albini_2021} showed that a mesh allows for implementing faster and simpler processing pipeline for feature extraction compared to the use of images.
Map structures are instead worth considering when a dimensionality reduction is required to ease the processing of data generated from taxels that may have a complex distribution. \\ 

Operational constraints can also influence the choice of the structure. For instance, choosing between images or point clouds for object recognition may depend on the object's size. Images can be used with small objects containing enough detailed features in a local area of interest. Larger objects instead generally require more than a single touch to be classified, thus requiring sharing multiple local information across different tactile images \cite{pezzementi_2011}. In this case, the whole shape of the object can be instead represented as a single point cloud and then classified using \gls{dl} methods \cite{qi2016pointnet}.

\subsubsection*{Multiple Representations}

The previously introduced \cref{img:overview}, where tactile object recognition is used as an example, shows a single point cloud representation supporting all subsequent stages of the pipeline. While this serves to illustrate the perception–action loop, it represents a conceptual case: in practice, complex tasks may require more than one data structure. As previously discussed at the end of Section~2, the Data Representation layer is not limited to a fixed, one-time transformation, but may change between stages to better fit task-specific requirements.

This is further illustrated in \cref{img:tasks}, which shows examples of complex tasks broken down into subsets of Processing, Inference, and Action operations. As seen in \cref{tb:repr_tasks}, no single representation can effectively support all of these operations. For example, in the case of Object Property Estimation (bottom-right of \cref{img:tasks}), a pipeline might include shape reconstruction, classification and texture recognition. Point clouds are suitable for shape reconstruction and classification, but they are not appropriate for texture classification, which would require a different structure. Consequently, the Data Representation layer should contain multiple structures representing the same underlying tactile data, allowing each processing stage to operate on the most suitable format.

Moreover, \cref{tb:repr_tasks} also shows that multiple structures can support the same operation. This is expected since, as discussed in Section~6, different structures can encode the same tactile information. While it may seem convenient to use a single representation across multiple operations within the same task, thereby avoiding the overhead of managing multiple data structures, practical considerations can justify the use of more than one format.
For instance, \cite{fan_2022} found that representing and processing the distribution of markers captured by a camera-based sensor as a mesh and using \gls{gcnns} significantly reduced training time compared to using images with \gls{cnns}. In such cases, it can be advantageous to employ images in the initial stages of processing (e.g., marker extraction, force computation) and then convert to a mesh for learning with \gls{gcnns}. Similarly, \cite{denei_2014} demonstrated a task in which contact location sensing and force control were supported by a mesh structure, while motion planning was carried out in a low-dimensional space represented as a map.

%\newpage
 
\section{Multi-Modal and Cross-Modal Data Representations}
\label{sec:multimodal}

\begin{table*}[]
	\centering
	\caption{Multi and Cross modal tactile data representations.}
	\setlength\extrarowheight{2pt}
	%\resizebox{\textwidth}{!}{
		\label{tb:multi}
		\begin{tabular}{cc|cccccc|}
		\cline{3-8}
		&                                                                                   & \multicolumn{6}{c|}{\textbf{Data Structure}}                                                                                                                                                                                                          \\ \cline{2-8} 
		\multicolumn{1}{c|}{}                                          & \textbf{\begin{tabular}[c]{@{}c@{}}Additional\\ Sensory Information\end{tabular}} & \multicolumn{1}{c|}{\textit{Vector}} & \multicolumn{1}{c|}{\textit{Matrix}} & \multicolumn{1}{c|}{\textit{\begin{tabular}[c]{@{}c@{}}Point\\ Cloud\end{tabular}}} & \multicolumn{1}{c|}{\textit{Mesh}} & \multicolumn{1}{c|}{\textit{Image}} & \textit{Map} \\ \hline
		\multicolumn{1}{|c|}{\multirow{5}{*}{\textbf{Multi-Modality}}} & \textit{Proximity}                                                                & \multicolumn{1}{c|}{$\CIRCLE$}      & \multicolumn{1}{c|}{}                & \multicolumn{1}{c|}{}                                                               & \multicolumn{1}{c|}{}              & \multicolumn{1}{c|}{$\Circle$}      &     $\Circle$         \\ \cline{2-8} 
		\multicolumn{1}{|c|}{}                                         & \textit{Vibration}                                                                & \multicolumn{1}{c|}{$\CIRCLE$}      & \multicolumn{1}{c|}{}                & \multicolumn{1}{c|}{}                                                               & \multicolumn{1}{c|}{}              & \multicolumn{1}{c|}{}               &              \\ \cline{2-8} 
		\multicolumn{1}{|c|}{}                                         & \textit{Temperature}                                                              & \multicolumn{1}{c|}{$\CIRCLE$}      & \multicolumn{1}{c|}{$\Circle$}       & \multicolumn{1}{c|}{}                                                               & \multicolumn{1}{c|}{}              & \multicolumn{1}{c|}{$\Circle$}               &              \\ \cline{2-8} 
		\multicolumn{1}{|c|}{}                                         & \textit{Color and Shape}                                                          &   \multicolumn{1}{c|}{$\Circle$}               & \multicolumn{1}{c|}{$\Circle$}       & \multicolumn{1}{c|}{}                                                               & \multicolumn{1}{c|}{}              & \multicolumn{1}{c|}{$\CIRCLE$}      &              \\ \cline{2-8} 
		\multicolumn{1}{|c|}{}                                         & \textit{Depth}                                                                    & \multicolumn{1}{c|}{}               & \multicolumn{1}{c|}{}                & \multicolumn{1}{c|}{$\CIRCLE$}                                                      & \multicolumn{1}{c|}{}              & \multicolumn{1}{c|}{}               &              \\ \hline
		\multicolumn{1}{|c|}{\multirow{2}{*}{\textbf{Cross-Modality}}} & \textit{Color and Shape}                                                          & \multicolumn{1}{c|}{$\Circle$}      & \multicolumn{1}{c|}{}                & \multicolumn{1}{c|}{}                                                               & \multicolumn{1}{c|}{}              & \multicolumn{1}{c|}{$\CIRCLE$}      &              \\ \cline{2-8} 
		\multicolumn{1}{|c|}{}                                         & \textit{Depth}                                                                    & \multicolumn{1}{c|}{}               & \multicolumn{1}{c|}{}                & \multicolumn{1}{c|}{$\CIRCLE$}                                                      & \multicolumn{1}{c|}{}              & \multicolumn{1}{c|}{$\CIRCLE$}      &              \\ \hline
	\end{tabular}
		%}
	\begin{tablenotes}
		\item[*] The table shows how pressure or force information is represented along with additional sensing modalities. Filled circles represent data structures used by the majority of works presented in the literature. Empty circles indicate alternatives considered by a significantly lower number of works.
	\end{tablenotes}
\end{table*}

So far the review has addressed the representation of tactile information consisting of forces or pressures as well as their distributions and locations. However, the sense of touch is inherently multimodal - from physical contacts additional information including temperature and vibrations can be captured. Furthermore, beyond sensing modalities related to tactile perception, in recent years there has been growing interest in combining tactile sensing with visual feedback. 
With respect to the architecture described in \cref{img:overview}, data fusion is mostly related to Processing and Inference operations requiring the extraction of information from multiple feedback.

This section analyzes how tactile data, particularly pressure or force, is represented alongside additional sensory inputs. The 
analysis is split into two main categories: multi- and cross- modality. The term \textit{multi-modality} refers to applications where sensors are combined to compensate for the lack of one modality with another. An immediate example of this can be found in the context of object classification \cite{liu_2017_1}, where cameras can rapidly capture the global information on the shape of an object, while a tactile sensor can capture local shape information along with mechanical properties. Combining the two modalities allows discriminating objects using both geometrical and mechanical information.
The second category refers to \textit{cross-modality} - i.e., different modalities do not directly interact but can be seamlessly interchanged. 
Again, using object recognition as an example, the cross-modal classifiers proposed in \cite{parsons_2022,falco_2019} allow for training a classification model on visual data only and deploying it for a task of tactile-based classification. 

The outcome of the analysis is shown in \cref{tb:multi}, reporting how data can be represented with additional sensing modalities. 
Each element in the table has the same meaning as Section 7
% \cref{sec:tasks} 
- filled circles indicate data structures used by the majority of the literature, while empty circles represent alternatives less commonly adopted. 
It is also important to note that \cref{tb:multi} does not explicitly include the robot’s kinematic information. This omission is intentional, as kinematics are typically implicitly embedded within data representations that encode the spatial locations of taxels, such as point clouds or mesh-based structures, which are typically calibrated relative to the robot’s body or base frame.

\subsection{Multi-Modality}

This section first examines how tactile sensing is represented when combined with low-spatial-resolution sensors typically embedded in the robot’s body. It then discusses integration with camera-based sensing, which is used to capture high-resolution, dense measurements from the environment.

\subsubsection*{Low Spatial Resolution Sensing}

This category includes sensors that provide coarse spatial information and are typically distributed with low resolution across the robot’s body. Examples include sensors providing temperature, vibration, and proximity information. 

A first class of multi-modal applications combining these modalities with tactile sensing focuses on analysing time series data in the context of textures \cite{kaboli_2018,hoelscher_2015,chathuranga_2013,chathuranga_2013_1}, gestures \cite{kaboli_2015} or object \cite{chu_2015,tanaka_2014,xu_2013} recognition.
As discussed in Section 7, sensor outputs are typically represented as vectors, which are then processed to compute descriptors used as input features for classification with \gls{ml} techniques.
Sensing modalities represented in this way are mainly related to vibrations \cite{kaboli_2018, kaboli_2015,hoelscher_2015,chathuranga_2013,chathuranga_2013_1,tanaka_2014,xu_2013} or temperature \cite{hoelscher_2015,chu_2015,xu_2013}. 
In a similar fashion, time series are used for tactile-proximity fusion in classification tasks \cite{palermo_2020} and for detecting transitions between pre-touch and contact phases \cite{rupavatharam_2023}.

Beyond vector-based representations, \cite{li_2020} proposed using matrices to organize multimodal sensor data. Their sensor integrates four modalities (pressure, object temperature, external temperature, and thermal conductivity), across ten sensor units mounted on a robotic hand. The matrix representation stacks the responses of different modalities over the rows. 
Another solution is to represent data as images when taxels provide multimodal information and are arranged as a regular grid.
Examples can be found in \cite{ge_2023,navarro_2020}, for both temperature and proximity sensing combined with pressure data.

Finally, a recent work by \cite{liao_2024} introduces a map representation implemented as a graph structure that implicitly encodes the topology of distributed proximity and pressure sensors. This approach has proven effective for object recognition and grasp stability estimation tasks.

\begin{figure}[t!]
	\centering
	\includegraphics[width=0.95\columnwidth]{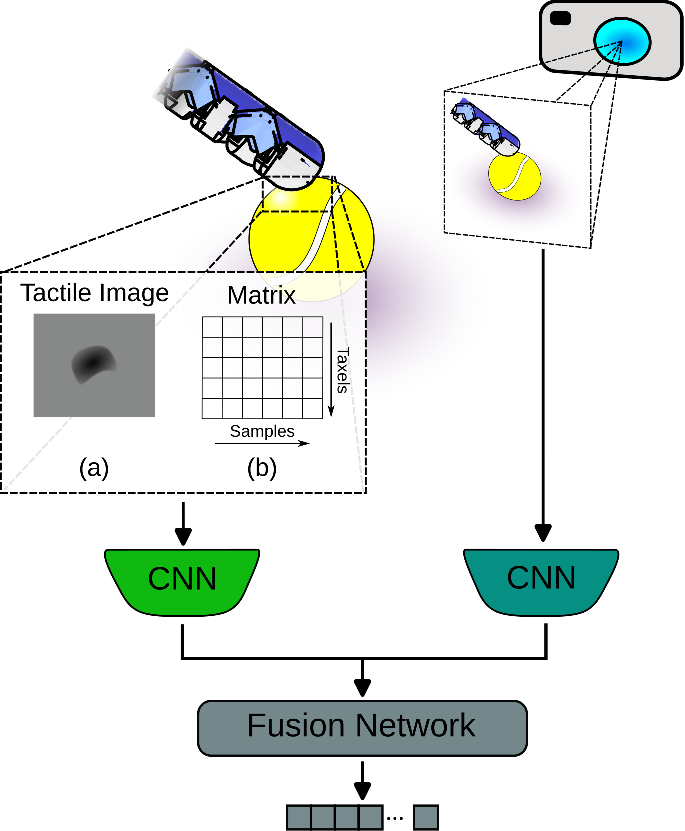}
	\caption{Common fusion strategy based on \gls{dl}. Visual images and tactile inputs are fed into separate \gls{cnns}. The network performing data fusion takes the output of each \gls{cnns}, which can be processed separately, concatenated or further elaborated by another network. Tactile inputs can consist of: (a) a tactile image encoding the local shape of the object at the contact location; (b) a matrix encoding temporal information of each taxel.
	}
	\label{img:visuo_tac}
\end{figure}

\subsubsection*{Visuo-Tactile Fusion}

Synergy between vision and touch is most commonly achieved by representing both tactile and visual data in the form of images  \cite{luo_2018,zhao_2023,pebdani_2023,luo_2015_1}. This clearly facilitates the application of established computer vision methods for both Processing and Inference operations.

A widely used fusion strategy involves feeding visual and tactile images into separate \gls{cnns} for feature extraction, followed by fusion at a higher network level. \cref{img:visuo_tac} illustrates the general architecture.
Examples can be found in the context of recognition \cite{luo_2018,lin_2019,pebdani_2023,shoujie_2023}, localization and/or shape reconstruction \cite{zhao_2023,bauza_2024,xu_2023,gao_2023,kerr_2022}, slip detection \cite{jianhuali_2018} or manipulation \cite{zhenyu_2023,sferrazza_2024,fu_2025,tong_2025}.
Transformer-based architectures have recently emerged as a powerful alternative for visuo-tactile integration. These models typically take tactile and visual images as input, (optionally applying a preprocessing step such as feature extraction) and feed the representations into the transformer for inference \cite{yunhai_2024,cui_2020,gu_2024}.

Earlier works not relying on \gls{dl} were based on the same underlying principle of processing tactile and visual images using classical image processing techniques. For example, \cite{luo_2015_1} processed tactile and visual data to localize the contact point within the visual image.

Another line of work transforms tactile time-series data into matrix-like representations.
These approaches attempt object recognition by reorganising data by stacking time series of tactile information. Then, visuo-tactile data is processed in the same way to look for correlations \cite{liu_2017_1,gao_2015,ding_2023,liang_2023,murali_2025}. For learning-based architectures, the process can be similar to the one previously described for tactile images and shown in \cref{img:visuo_tac}. 
In contrast, \cite{ueda_2024} demonstrates a multimodal architecture that combines visuo-tactile data where tactile signals are represented as vectors of time series, rather than reshaped images. Similarly, \cite{pugach_2019} employs vector representations to learn visual-motor associations.

\subsubsection*{Depth Integration}

In addition to high-resolution color and shape cues from standard cameras, depth information, acquired using RGB-D or \gls{tof} sensors, has been extensively leveraged alongside tactile sensing.
Depth-tactile fusion is particularly valuable for tasks involving control, manipulation, environment mapping, and object shape reconstruction~\cite{yuan_2023,ilonen_2014,murali_2021,murali_2022,caroleo_2024,dutta_2024,dutta_2025}.

In applications related to object geometry reconstruction, a visual point cloud provides a coarse estimate of an object’s geometry, which is incrementally refined by tactile interactions. Each touch yields a local point cloud capturing fine surface details, which is then merged with the visual data to improve shape accuracy and resolve occlusions.
To obtain smoother or continuous surfaces, point clouds may be further refined into triangular meshes \cite{smith20203d,smith2021active}, or interpolated to build a continuous representation on top of them \cite{rustler_2022,wang_2018,bjorkman_2013,suresh_2021_1}.  

Object representations created by combining tactile and depth data are typically limited to geometric information. These works often overlook mechanical properties that can be inferred through touch, such as friction or compliance.
To address this limitation, some works have proposed encoding physical properties directly into geometric representations. In particular, \cite{rosales_2014,nguyen_2021,yao_2023,caroleo_2025} process taxel measurements collected during contact to generate point clouds that not only reflect object geometry but also capture spatial variations in physical properties across the surface.
A common strategy is to color the point cloud to represent these properties: for instance, using color gradients to indicate local friction coefficients \cite{rosales_2014,nguyen_2021} or stiffness estimates \cite{yao_2023,caroleo_2025}. This results in a rich, multimodal surface representation that integrates both shape and touch-derived physical characteristics.

A different use of point clouds is presented in \cite{roncone_2015,roncone_2016,rozlivek_2025}, where a receptive field is associated with each taxel in Cartesian space to detect and predict tactile activations triggered by visual events occurring in the robot’s immediate proximity. 

\subsection{Cross-Modality}

The main motivation for cross-modality in tactile sensing is to reduce the burden of tactile data collection, which remains a significant bottleneck in training data-driven models.
Collecting large-scale tactile datasets is particularly challenging. Tactile sensors provide only local contact information. Therefore, tasks that require global understanding, such as recognizing large objects, demand numerous interactions with the environment \cite{pezzementi_2011}. Moreover, tactile responses are inherently action-dependent. As a matter of fact, to properly collect a dataset, objects must be explored multiple times by applying different forces or motions. 

Such data collection cannot be performed manually at scale. It requires robotic or automated systems, where motion planning must account for both kinematic constraints and the geometries of the objects and sensors involved.
These challenges have led to two main lines of cross-modal research: (i) using visual data to train models for tactile-based object recognition, and (ii) generating tactile outputs directly from visual inputs, leveraging \gls{dl} generative techniques.

\subsubsection*{Cross-Modal Object Recognition}

Tactile object recognition is a representative task that typically demands large volumes of training data. 
To reduce the extensive acquisition of tactile data, recent research has explored cross-modal learning approaches, where models are trained with data acquired using a different modality (usually easier to collect). 

Notably, \cite{falco_2019,parsons_2022,murali_2022,murali_2022_1} showed the benefit of exploiting the readiness and availability of visual data to train a system that can recognize objects through touch.
A key challenge in cross-modal object recognition lies in finding a shared representation that bridges the gap between the two sensing modalities. In this regard, point clouds have emerged as an effective data structure, providing a common format through which visual and tactile data can be encoded and processed in a unified way \cite{falco_2019,parsons_2022,murali_2022,murali_2022_1}.

\begin{figure}[t!]
	\centering
	\includegraphics[width=0.48\textwidth]{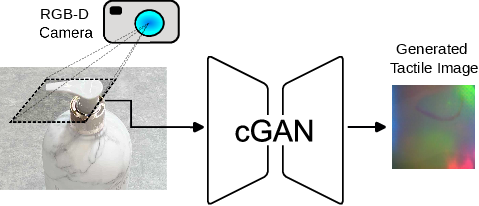}
	\caption{
		This figure illustrates the core concept behind cross-modal tactile data generation. A Conditional Generative Adversarial Network (cGAN) is trained to generate a tactile image from visual inputs, typically RGB and depth. The visual image isolates a surface feature of interest, and the network predicts the tactile response that would have been obtained with the robot physically interacting with a tactile sensor.
	}
	\label{img:cross_data_gen}
\end{figure}

\subsubsection*{Cross-Modal Tactile Data Generation}

Beyond object recognition, cross-modality has recently been explored as a means to generate tactile sensor outputs from visual input, particularly camera images
This line of work leverages generative models originally developed for computer vision~\cite{pan_2019}, adapting them for cross-modal tactile data generation and demonstrating their potential to mitigate some of the challenges associated with collecting tactile data.

The general strategy involves adapting image-to-image translation frameworks, such as \texttt{pix2pix} \cite{aziz_2020}, to convert visual inputs (e.g., RGB or depth images) into corresponding tactile data. \cref{img:cross_data_gen} illustrates this concept. These models are typically trained on paired datasets, where each visual sample is aligned with a tactile observation corresponding to the same geometric feature. 
Several studies have proposed methods for generating tactile data from RGB \cite{li_2019,cai_2021,lee_2019} or depth data \cite{patel_2020,zhong_2023,zhong_2024}. While there may be differences in terms of architecture or training pipeline, all of them encode tactile data as images. This is a direct consequence of employing models for image-to-image translation. 
A slightly different approach is proposed by \cite{takahashi_2019}, who employs an encoder–decoder architecture to generate vectors representing tactile time series from visual input. This strategy allows for representing the temporal dynamics of tactile interactions, moving beyond static image-based predictions.

\subsection{Discussion}

\cref{tb:multi} summarizes the previous analysis. Out of the six representations introduced in Section 5, %\cref{sec:build}, 
only three, vectors, point clouds and images, have been mainly used in the literature for multi-sensor data fusion. 

For sensors providing data with low spatial resolution such as proximity, vibration or temperature, two different approaches can be identified.
When data (in addition to tactile) is collected from different sensors \cite{palermo_2020,rupavatharam_2023,kaboli_2018, kaboli_2015,hoelscher_2015,chathuranga_2013,chathuranga_2013_1,chu_2015,tanaka_2014,xu_2013}, their number, positions and the sampling time of the measurements do not match the corresponding tactile part.
Therefore, vector structures are mainly exploited in this case since they
are agnostic from a geometric point of view. Furthermore, their lengths can vary to encode lower or higher-density information collected in a given time window.
More structured representations such as images or matrices are applied only when the multimodal data is generated from the same sensor \cite{navarro_2020,li_2020}. 
For example, the capacitive-based sensor in \cite{navarro_2020} is not shielded, thus providing both pressure and proximity information. Similarly, the sensor designed in \cite{li_2020} can provide four different modalities. In such a condition, measurements related to the different sensing modalities have the same spatial resolution and distribution. % 

When combining tactile sensors with cameras providing higher-density and structured data, it becomes natural to use images or matrices. A different density or resolution is not critical since images can easily be resized. Furthermore, nowadays high-level computation can be done with \gls{cnns} \cite{luo_2018,zhao_2023,pebdani_2023} - as long as the two inputs are represented as a grid (thus considering both images and matrices) they can be processed in the same way. In general, grid-like structures are the simplest and most obvious representation to merge visuo-tactile data.

Tactile data is instead mainly represented in the form of point clouds when paired with depth information. Small tactile point clouds are collected for each different contact and merged with camera depth data to obtain point cloud-based representation of the whole object or environment. 
In this respect, the authors want to highlight the fact that, although a lot of effort has been put into modelling the environment surrounding the robot, tactile data has been generally used to encode geometric features. 
While the sense of touch has the potential to offer more, when combined with vision it is predominantly utilized akin to high-resolution cameras, providing a detailed reconstruction of the local surface of interest. 
Very few works effectively use taxel measurements with the goal of obtaining a representation of objects also encoding their mechanical properties \cite{rosales_2014,nguyen_2021,yao_2023,caroleo_2025}.

It is also worth noting that, as visible from \cref{tb:multi}, meshes are not currently an option for multi-modal tasks, while maps have been considered only by \cite{liao_2024}. %\TODO{update with \cite{pitti_2017}}
As explained in Sections 5 and 7, % \cref{sec:build} and \cref{sec:tasks},
they are useful to explicitly or implicitly represent complex or sparse taxel distributions. 
Multi-modal applications working with sparse data are more targeted to object classification, localisation, or shape reconstruction, usually performed with small-area sensors embedded into fingertips, and whose Processing and Inference methods are inherited from computer vision techniques mostly based on images and point clouds. 

Research on Cross-Modality explores how to learn representations of the objects using complementary sensory information. 
Current cross-modal applications are mostly related to two tasks - object recognition and tactile data generation. Since these applications are linked to computer vision problems, it is not surprising that mainly point clouds and images are used. 
In particular, generative models for tactile data have mostly been applied by modelling both visual and tactile data as images. However, we argue that future research will explore alternative data representations. For example, generative models for mesh structures have already been proposed \cite{mansimov_2019,samanta_2020,macedo_2024} and may soon be adapted for tactile data generation. 
Another promising direction on this subject is also given by \gls{fms}. Nowadays, they have been used for cross-modal data generation, to convert images or videos and generate textual descriptions (and vice-versa). More recently, they have begun to be applied to touch understanding, cross-modal retrieval, and image generation using tactile images as input~\cite{yang_2024}. However, their current use remains limited to image-based tactile inputs and they still require large amounts of data for training.

\section{Conclusions}
\label{sec:conclusion}

Robotic tactile perception is complex and implemented through a set of operations related to Action, Data Acquisition, Data Representation, Processing and Inference (see \cref{img:overview}). Each of these impacts the overall perception process, shaping tactile information. This review specifically focused on Data Representation, which is required to interface raw data provided by tactile receptors distributed on the robot's body with high-level processing and computation methods.
The literature has been reviewed to understand how tactile information has been represented by previous works.
From this analysis, six data structures commonly used in the literature have been identified and general guidelines have been provided to select them depending on (i) the available hardware, (ii) the type of information encoded, and (iii) the operations required by the task. The discussion is also extended in Section 8
%\cref{sec:multimodal} 
to consider the use of additional sensing modalities. 
Key aspects and takeaways related to the main points discussed in the review are highlighted in \cref{tb:summary}. 

\newcolumntype{L}[1]{>{\raggedright\arraybackslash}p{#1}}
\begin{table*}[t]
	\renewcommand{\arraystretch}{1.2}
	\centering
	\caption{Summary of key insights and implications from the previous sections.}
	\resizebox{\textwidth}{!}{
	\label{tb:summary}
	\begin{tabular}{|L{2.2cm}|p{7.0cm}|p{7.0cm}|}
		\hline
		\textbf{Aspect} & \textbf{Key Insights} & \textbf{Implications} \\
		\hline
		\textit{Role of Data Representation} & Tactile perception is an action-conditioned process involving different steps. Data Representation allows for structuring data into formats suitable for high-level processing & It serves as the bridge between sensor hardware and computation. The chosen data structure influences all subsequent steps in the tactile perception pipeline. \\
		\hline
		\textit{Hardware Abstraction} & A generic tactile sensor can be seen as composed of taxels with different spatial distributions. & Tactile data can be represented with any of the structures presented in Section~5, by transforming the spatial arrangement of the taxels. % Representations can generalize across different sensing technologies A mismatch between the data structure and the taxels’ spatial distribution can lead to loss of information. %Representation choice should be guided by task requirements rather than hardware alone. 
		\\
		\hline
		\textit{Effects on Tactile Information} & Representing with a given structure implicitly filters tactile information. Spatial properties of the taxels may not be fully preserved if their layout is incompatible with the chosen structure. & Many tasks use only a fraction of the available tactile information. Representation should be selected with prior knowledge of the specific contact information required for the task. \\
		\hline
		\textit{Application-based Selection} & No single representation can support all the operations listed in \cref{tb:repr_tasks}, though some can be applied to multiple applications.%Vectors and matrices are mainly used for tasks involving contact dynamics; images and point clouds are preferred for representing and processing spatial information; meshes and maps are primarily used for large or curved taxel distributions. 
		& Complex applications may require using multiple representations of the same raw tactile data at different stages of the perception pipeline. \\
		\hline
		\textit{Multi and Cross Modality} & Data Representation enables seamless integration of touch with other modalities. Vectors are often used to fuse tactile data with low-resolution sensors, where spatial information is largely neglected. Integration with dense measurements typically uses image or point cloud structures. & When geometric features are included, tactile sensing is often treated similarly to high-resolution cameras, overlooking unique tactile properties. %Richer fusion with vibration, temperature, and proprioception remains underexplored. 
		\\
		\hline
	\end{tabular}
    }
\end{table*}

In general, the authors' advice is to prioritise the computation performed at a high level. Hardware-related aspects are important but less critical. Indeed, as described in Section~5,  % \cref{sec:build}, 
data structures can be created from different hardware. 
Conversely, there is no single representation that fits all the possible operations (see \cref{tb:repr_tasks}). The hardware is worth considering when the same task can be accomplished with more than a single representation, as explained at the end of Section 7. % \cref{sec:tasks}. 
Clearly, this is valid from the point of view of the representation only, assuming that the available hardware is properly selected for the specific application~\cite{dahiya_2010}. \\ 

%%%%%%%%%%%%%%
The majority of current and past literature has focused on applications involving contact over small and flat areas, occurring at fingertips, grippers, or on rigid surfaces. Coupled with rapid advances in \gls{dl} methods for computer vision, this has led to a preference for certain types of representations that treat tactile sensing as analogous to image-based vision.
Although a few notable examples of hardware for whole-body tactile sensing exist, their high cost and system complexity have limited their use to only a few research groups worldwide. As a result, representations suitable for encoding contact data distributed over large areas (such as maps or meshes) remain underutilized.
However, the current trends related to soft robotics and humanoids suggest the need for algorithms capable of handling contacts distributed over large, soft surfaces that embed various types of transducers.\\

\subsection{The Increasing Importance of Data Representation}
	
The robotics market is rapidly expanding, with growing investment from both startups and leading companies, particularly in the area of humanoid development. These efforts aim to deploy robots in industrial settings within the next few years, and gradually integrate them into homes to assist with daily tasks within the next decade. This marks a significant shift from the traditional paradigm, where robots operated in isolated environments performing repetitive, pre-programmed tasks.
	
While the vision of robots as human-assistive systems is not recent, it is only now becoming feasible thanks to technological advances, cost reductions, and breakthroughs in Artificial Intelligence. However, building general-purpose robots that operate in unstructured environments and collaborate with humans remains highly challenging. Today’s systems mostly rely on collision avoidance - cameras or lidars are used to prevent contact, and robots halt operation when a collision occurs. Physical interaction is largely restricted to fine manipulation tasks. This paradigm is overly conservative and ill-suited for human-robot shared spaces.

To move beyond this limitation, robots must be equipped with sensitive bodies that support high-level reasoning based on tactile feedback. Compliance and body softness are expected to improve safety aspects, while whole-body sensing will be required for tasks such as bi-manual or whole-body manipulation, collision detection and reaction, and the interpretation of physical gestures. In this evolving landscape, the role of the data representation layer becomes increasingly important and its future relevance can be analysed from two complementary perspectives discussed in Section 3.

\subsubsection*{Hardware Perspective} The majority of recent works are largely based on the use of camera-based tactile sensors \cite{zhang_2022}. Their introduction significantly contributed to the growth of the tactile sensing community, as they are cheap, easy to manufacture and provide high spatial resolution, making them suitable for applications requiring object recognition or fine manipulation.
As a matter of fact, the concept of hardware abstraction is often overlooked by researchers using these sensors as tactile data is natively structured as an image.
However, its importance must not be underestimated. In the vision of general-purpose robots, the role of the sense of touch goes far beyond the need for high-spatial resolution. Furthermore, these sensors have limitations in applications where high bandwidth or hardware capable of scaling to cover the entire robot body and of capturing large contact distributions are required. Despite some promising developments \cite{duong_2021,funk_2024,luu_2025}, scaling these technologies to cover the robot body remains a significant challenge.
A more realistic scenario involves heterogeneous tactile coverage - soft and flexible sensors on curved surfaces, stretchable sensors at joints, and high-resolution devices on fingers.
In such a scenario, a robust data representation layer is essential to abstract hardware differences from the task. 
	
\subsubsection*{Computational Perspective}
This need for abstraction is equally critical on the computational side. General-purpose robots must perform a wide range of tasks, addressed using diverse computational tools. This requires structured, meaningful input representations.
Past advancements in tactile data processing and high-level reasoning have been mostly driven by breakthroughs and achievements in other areas of robotics or computer science. This has led to the improvement of existing applications (such as object recognition) and enabled new research areas (e.g., tactile data generation). 
It is reasonable to assume that this trend will not change in the near future - progress in other domains will be transferred to robotics and this will require data properly structured to enable the use of new computational techniques.
An example is the recent introduction of \gls{llms} and Transformers. Originally developed for Natural Language Processing \cite{patwardhan_2023}, they are now revolutionizing the robotics landscape, being used for human-robot interaction, task planning, visual perception, decision making, learning from demonstration and multimodal integration. Furthermore, as discussed in Section 8, %\cref{sec:multimodal}, 
they have been used to learn task representations by fusing visuo-tactile data. 
There will be a need to handle the unique characteristics of tactile data, which is inherently multimodal and spatially distributed over complex geometries of the robot body. This  will likely involve the use of different specialized data representations, each tailored to highlight one or more aspects of tactile information.

\subsection{Research Gaps and Future Directions}

In light of the increasing importance of tactile data representation for general-purpose robotics, the authors now identify current gaps and specific research directions that must be addressed to realize this vision.
Furthermore, as tactile applications become more complex and embodied, this demands moving from static structures to algorithms or learned representations capable of encoding sensory-motor contingencies. \\

\subsubsection*{Multimodal Middleware} 
A crucial focus for the community must be related to the development of a robust multimodal middleware that supports data collection from diverse sensors while ensuring time-synchronization, thus providing consistent data across modalities. Previous work on this subject has been done by \cite{youssefi_2015} for tactile sensors, but it lacks support for multimodal perception. 
From a technical point of view the Data Representation block can be implemented as part of the middleware that
manages the creation and update of the data structures over time, exposing them to the software implementing high-level processing operations. This will favor code reuse and portability to different hardware platforms, as the processing performed by the software can be applied to different sensors as long as they are represented with the same data structure.

\subsubsection*{Tactile Data Representation for Soft Robots}

Robots with soft bodies are expected to play a fundamental role in the near future due to their inherent safety in physical interactions. However, most current tactile sensing technologies embed taxels in rigid components of the robot. In contrast, soft robotic applications often demand sensors that can be integrated into compliant, deformable surfaces.
While hardware solutions have already been proposed to address this need, significant challenges remain on the side of data representation, as most existing methods assume fixed, rigid sensor layouts.

As discussed in Section 4, the extended definition of taxels includes marker-based sensing systems, such as those tracked via camera, which effectively behave as soft, deformable tactile elements. Prior work has demonstrated that meshes can serve as a suitable representation for encoding both topological and spatial relationships in these non-rigid settings \cite{fan_2022,quan_2023}. Specifically, when taxel positions vary over time (e.g., due to deformation), meshes can capture their evolving configuration and even the overall deformation of the robot's surface.
In this respect, meshes as a representation for tactile perception in soft robotics, are promising as they allow managing non-rigid sensor distributions. Nonetheless, further work is needed to develop computational tools and learning methods that can robustly model large sensor deformations in real-time.

\subsubsection*{Kinematics-Dependent Map Representations}
As introduced in Section 5, tactile maps offer a way to represent the robot body by projecting complex, non-planar geometries onto a lower-dimensional domain. Current mapping techniques typically preserve either spatial or topological relationships but are generally static and fixed.

For example, the maps proposed in \cite{denei_2014,sutanto_2019} are constructed using tactile sensors integrated into a single rigid part of the robot. This limits scenarios where tactile information must be processed across multiple links. To extend this approach, the mapping function would need to incorporate relative transformations between taxels distributed on different links, which are inherently dependent on the robot’s posture. \\
Other approaches, such as \cite{hoffmann_2018,brock_2009,pugach_2016}, instead allow the correlation of responses from sensors integrated into articulated bodies. However, these methods rely on a fixed structure where topological relationships do not adapt to changes in joint configurations or kinematics. As a result, the topological organization of the tactile map becomes misaligned with the actual physical configuration of the robot during motion. Addressing this limitation requires developing kinematics-aware mapping methods that dynamically update spatial or topological relationships based on the robot’s configuration.

\subsubsection*{Zero-Shot Transfer Across Robot Embodiments} 

Applications such as object identification, geometric feature recognition, and gesture recognition are typically supported by \gls{ml} methods that require substantial amounts of data. As previously noted, collecting tactile data is both time-consuming and sensor-specific. Moreover, due to hardware differences among sensing technologies, an ML model trained on data from one sensor is unlikely to generalize well to a different tactile sensor, due to differences in transduction mechanisms and the spatial distribution of taxels.\footnote{Although this limitation also applies to model-based approaches, it is generally less critical, as such methods usually require tuning rather than  a completely new data collection.}
While some recent works have addressed this issue by learning mappings between the outputs of two sensors for the same input~\cite{rodriguez2024, grella2024}, these approaches remain largely limited to finger-shaped tactile sensors and small-area contacts, and still require substantial amounts of data to train the mapping model.

Furthermore, beyond differences in transduction technologies, this challenge also arises when transferring algorithms between two sensors based on the same underlying technology but with different taxel distributions. For example, in whole-body sensing, the local distribution of taxels may vary due to differences in density or surface curvature, or between different robot links. As a result, a \gls{ml} model that relies on spatial features and is trained on a specific area may fail when applied to another part of the robot’s body.
For instance, a classifier trained to recognize gestures on a flat robot surface may not perform well on a rounded body segment, even if the gesture is physically identical. Retraining or fine-tuning the classifier for each body segment would require new data collection and is not a scalable solution.

This case highlights a broader challenge: the need for data representations that are invariant to the geometry of the sensor layout, while still preserving the structure of spatial contact information. In this way, the generalization problem is shifted from data collection to data representation, offering a more scalable solution. A possible direction is to develop tactile representations that abstract from the specific spatial arrangement of taxels and instead capture high-level, transferable descriptions of contact geometry. Such representations would decouple the encoded information from the sensing surface geometry, enabling zero-shot transfer of a classifier to another robot.

\subsubsection*{Transformers as Data Representation} 

Transformer-based architectures have gained traction in robotics for tasks involving multimodal processing, particularly in combining vision and touch to extract task-relevant features. Beyond their role in Processing and Inference, we argue that Transformers may also play a fundamental role in tactile data representation itself.
At the core of Transformers lies the self-attention mechanism, which weighs inputs based on their relevance to a given input. 
In tactile systems, it could be similarly applied to weigh sensor activations in different areas of the robot body depending on the current task, thus creating an implicit topological model akin to the artificial somatosensory maps defined in Section 5.
More broadly, Transformer encoders could serve as the backbone of fully end-to-end tactile perception architectures - performing not only processing, inference, and action selection, but also the representation of tactile data itself.

A recent paper showing this promising direction is presented by \cite{higuera_2024}, where Transformer-based models are trained on tactile images to build a reusable representation of touch across multiple tasks. However, as these models operate on tactile images, they inherit the strengths and limitations of image-based representations discussed in Section 6.
While highly effective for processing small and planar contact areas, their applicability to non-planar or whole-body tactile sensing remains uncertain.

Future research should then explore whether alternative data structures, such as mesh-based or topology-aware representations, enable better generalization for Transformer-based tactile architectures, especially in scenarios involving complex or large-area contact distributions. Another direction involves bypassing predefined spatial structures - feeding raw taxel data directly to Transformer models and allowing them to learn spatial and topological relations implicitly.

One of the primary challenges of applying Transformers to tactile perception lies in the sheer volume of data required for training. Transformers are data-hungry models, and currently, there are few publicly available datasets of tactile data, and they are highly sensor-specific. While techniques like transfer learning or fine-tuning on pre-trained models can help, the specific nature of tactile data, which is different from text or vision, would still require a substantial amount of task-specific data.

\subsubsection*{Representation of Multi/Cross-Modal Data and Robot Peripersonal Space}

As discussed in Section 8, tactile sensing is often combined with vision and represented using formats such as images or point clouds, primarily to leverage machine learning tools originally developed for computer vision. These representations are particularly effective at capturing the spatial structure of tactile data. However, they often overlook other important aspects of tactile perception. 
Similarly, temporal information is rarely integrated in a meaningful way. Even more underutilized is the encoding of motor actions, despite the fact that tactile signals are action-dependent. While existing data structures can include time as an extra dimension, they generally lack the capacity to encode sensory-motor contingencies explicitly or implicitly.
Modelling or learning these contingencies, i.e., predicting tactile feedback based on motor commands, would offer a more embodied and predictive form of perception. 

To fully capture interaction with the environment, such models also require visual information about the robot’s surroundings. Representations that integrate both the body and the nearby space are often referred to in robotics as \gls{pps}, typically defined as the region immediately surrounding the body, within reachable distance.
Previous work on \gls{pps} representations has primarily focused on fusing visual and proprioceptive signals. While some approaches have incorporated tactile feedback \cite{hikita_2008,copete_2016,pugach_2019,nguyen_2019}, the tactile modality is usually reduced to binary contact detection or used indirectly to improve visual-motor models. In contrast, \cite{roncone_2016} presents an approach in which tactile data plays a central role in building a visuo-tactile \gls{pps}.

In light of recent advances, the authors believe that extensions of multimodal \gls{llms} and \gls{vla} models represent a promising direction for developing embodied representations that integrate visual, tactile, and motor information. These models have the potential to learn high-level abstractions of physical interaction, capturing not just sensory data, but their evolution in time and their relationship to motor actions in a task-agnostic manner.

From this perspective, the concept of tactile data representation may evolve beyond fixed data structures toward learned representations or algorithms that directly encode sensorimotor contingencies. Such models could predict tactile responses from visual input, anticipate future actions or sensory states based on visuo-tactile sequences, and compress multimodal data (and their temporal evolution) into compact latent representations that generalize across different tasks and robot morphologies.
This moves data representation from being a structuring layer to becoming an active, predictive component for perception and control.

\section*{Appendix: Spatial Calibration for Sensors Distributed Over a Large Area}
\label{app:spatial_calib}

As previously seen, some representations are built from knowledge of the taxel positions or orientations with respect to a common reference frame.
If this information is available, the tactile sensing system is \textit{spatially calibrated}. 
As discussed at the end of Section 5, the problem of finding the position and orientation of tactile elements becomes particularly hard for technologies covering a large part of the robot and composed of a high number of taxels, such as those presented by \cite{ohmura_2006,cheung_1989,minato_2007,maiolino_2011,mittendorfer_2011}. 

Although it seems that the position of tactile elements can be retrieved from CAD when designing the placement of the sensors, two aspects must be considered: (i) large-area tactile sensors are usually manufactured as flexible and planar PCBs, which have to be cut and adapted to curved surfaces. Even with the support of CAD files, it is hard to exactly predict how elements belonging to a 2D flexible geometry will be redistributed on an arbitrary non-planar surface; (ii) the actual integration is performed manually, thus adding uncertainties - the real placement of the taxels may differ from the desired one. Furthermore, even the integration of sensors into the same robot part may differ among multiple batches. 

An accurate manual calibration of possibly a thousand tactile elements integrated into a complex robot body is an infeasible task. 
Therefore, semi-autonomous procedures to achieve spatial calibration have been investigated in the literature \cite{cannata_2010,prete_2011,mittendorfer_2012,mittendorfer_2014,albini_2017_1,rustler_2020}. 
These methods process the output of different sensors, such as force/torque or cameras, to estimate the position and orientation of taxels.
It is interesting to note that these works are based on different assumptions that make the methods valid under specific conditions, which are not generally valid for all the different tactile sensing hardware. As a result, fully autonomous methods designed to spatially calibrate different types of large-area tactile system are still missing. 

In particular, the aforementioned works may require: 
(i) the knowledge of a contact model \cite{cannata_2010}; 
(ii) the knowledge of the robot body through CAD files \cite{prete_2011,rustler_2020}; 
(iii) a specific setup or equipment \cite{cannata_2010,prete_2011};
(iv) the knowledge of the response model of the taxels \cite{albini_2017_1}.
Furthermore, they may be specifically designed for a given tactile system \cite{mittendorfer_2012,mittendorfer_2014} or for a class of tactile sensors  \cite{albini_2017_1}.

To evaluate the quality of the resulting calibration, the error between the true and estimated locations of the taxels is usually computed and averaged. In this respect, two types of errors can be generally computed. The first is the absolute error that averages the element-wise distance between each pair of real and predicted taxel positions and orientations \cite{cannata_2010,mittendorfer_2014,albini_2017_1}. The second is the relative error, which evaluates how well the geometric relations are preserved (see Section 6.3) with respect to the real placement \cite{prete_2011,mittendorfer_2014,albini_2017_1,rustler_2020}. Compared to the absolute error, this is computed in a similar way but first subtracting the centroid of the taxel distribution from both the real and estimated placements. \\
It is important to remark that an exact estimation of the error can be achieved only in simulation \cite{cannata_2010} or when the taxels distribution lies on a plane and the geometric model of the patches is available. Indeed, in the non-planar case, the real placement is in general not available and the error metrics are computed with small uncertainties. In these cases, the calibration procedure can be validated by integrating sensors into a simple 3D object, such as a cylinder, where the real taxel placement can be manually calibrated \cite{mittendorfer_2012,mittendorfer_2014}. Alternative solutions exploit a camera to identify the relative positions of the taxel (neglecting camera-related inaccuracies) \cite{rustler_2020} or evaluate the error on a plane \cite{albini_2017_1}.

These relative and absolute errors are generally used to assess the performance of the calibration method. However, when comparing different approaches, the hardware used for validation should also be considered. As an example, a 5 millimetre calibration error is acceptable if the relative distance among adjacent taxels (i.e. the pitch) is much larger than 5 millimetres. On the contrary, the uncertainty in the taxels position would be larger than the real relative pitch. 
In general, we argue that the outcome of a semi-autonomous spatial calibration procedure can be considered acceptable when the absolute or relative errors are \textit{lower} than the pitch between taxels. To the best of our knowledge, this was achieved by \cite{albini_2017_1} and \cite{rustler_2020} only.

	%\begin{acks}
	%	This work was supported by the SESTOSENSO project (HORIZON EUROPE Research and Innovation Actions under GA number 101070310).
	%\end{acks}
	
	%\bibliographystyle{SageH}
	\bibliographystyle{unsrt}
	\bibliography{ref}
			
\end{document}